\pdfoutput=1
\documentclass[11pt]{article}

\usepackage{acl}

\usepackage{times}
\usepackage{latexsym}

\usepackage[T1]{fontenc}
\usepackage[utf8]{inputenc}

\usepackage{microtype}

\usepackage{inconsolata}

\usepackage{amsthm}
\usepackage{amsmath}
\usepackage{amssymb}
\usepackage{booktabs}
\usepackage{bm}
\usepackage{tikz}
\usepackage{graphicx}
\usepackage{subfig}
\usepackage{algorithm}
\usepackage{algpseudocode}
\usepackage{listings}
\usepackage{caption}
\usepackage{makecell}
\usepackage{multirow}

\usepackage{appendix}

\usepackage{xspace}

\makeatletter
\DeclareRobustCommand\onedot{\futurelet\@let@token\@onedot}
\def\@onedot{\ifx\@let@token.\else.\null\fi\xspace}

\def\eg{\emph{e.g}\onedot}

\def\ie{\emph{i.e}\onedot}

\makeatletter
\def\@fnsymbol#1{\ensuremath{\ifcase#1\or \dagger\or \ddagger\or
   \mathsection\or \mathparagraph\or \|\or **\or \dagger\dagger
   \or \ddagger\ddagger \else\@ctrerr\fi}}
\makeatother

\title{AdaRefiner: Refining Decisions of Language Models with\\ Adaptive Feedback}

\author{Wanpeng Zhang \\
  School of Computer Science \\
  Peking University \\
  \texttt{wpzhang@stu.pku.edu.cn} \\\And
  Zongqing Lu\thanks{Corresponding Author} \\
  Peking University \\
  BAAI \\
  \texttt{zongqing.lu@pku.edu.cn} \\}

\begin{document}
\maketitle

\begin{abstract}
Large Language Models (LLMs) have demonstrated significant success across various domains. However, their application in complex decision-making tasks frequently necessitates intricate prompt engineering or fine-tuning, leading to challenges in unseen downstream tasks and heavy demands on computational resources. Meanwhile, Reinforcement Learning (RL) has been recognized as effective in decision-making problems but struggles in environments with sparse rewards, such as open-world games. To overcome these challenges, we introduce AdaRefiner, a novel framework designed to enhance the synergy between LLMs and RL feedback. The key component of AdaRefiner is a lightweight Adapter Language Model (LM), which automatically refines task comprehension based on feedback from RL agents. This method mitigates the need for intricate prompt engineering and intensive LLM fine-tuning while maintaining the LLMs' generalization abilities and enhancing their decision-making capabilities in downstream tasks. Empirical evaluations of AdaRefiner on $22$ diverse tasks within the open-world game \textit{Crafter} have demonstrated its superior effectiveness, especially in guiding agents towards higher-level and common-sense skills. Our work makes contributions to the automatic self-refinement of LLMs with RL feedback, offering a more adaptable and efficient solution for complex decision-making problems. The code is available at \url{https://github.com/PKU-RL/AdaRefiner}.
\end{abstract}

\section{Introduction}

\begin{figure}[t]
\centering
\includegraphics[width=\columnwidth]{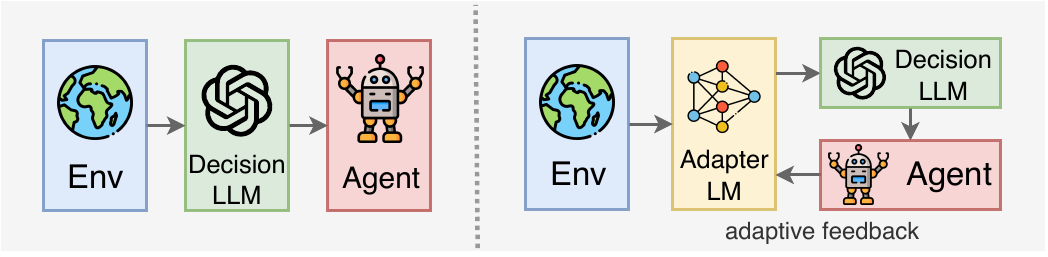}
\caption{Core differences between AdaRefiner (\textit{right}) and typical LLM-based methods (\textit{left}). The key distinction is the integration of Adapter LM, which enhances the synergy between LLMs and adaptive feedback.}
\label{fig:pipeline}
\end{figure}

The rapid development of Large Language Models (LLMs), trained on massive corpora, has opened new frontiers in various fields, leveraging their ability to process and generate text \citep{wei2022emergent}. Notably, LLMs have demonstrated impressive performance in decision-making problems \citep{yao2023react, shinn2023reflexion, sun2023adaplanner}. However, recent studies highlight that directly applying LLMs to complex decision-making tasks often necessitates intricate prompt engineering and external feedback \citep{wang2023voyager, wu2023spring, wang2023mint}. Such task-specific designs pose challenges in transferring these methods to different scenarios. Some studies have explored the use of task-related data to fine-tune LLMs to improve decision-making capabilities \citep{nottingham2023embodied,feng2023llama}. However, such approaches often encounter practical challenges, such as inaccessible LLM weights or intensive computational demands. Moreover, fine-tuning LLMs may lead to decreases in their generalization capabilities \citep{wang2022preserving}, making their deployment across diverse environments challenging. These challenges underscore the need for a more adaptable and generalizable approach.

Before the emergence of LLMs, Reinforcement Learning had been recognized for its impressive capabilities in decision-making problems \citep{mnih2015human,silver2017mastering}. The strength of RL is most evident when agents consistently receive clear and dense rewards that guide them toward the targeted behaviors \citep{ladosz2022exploration, eschmann2021reward}. However, designing such reward functions is far from straightforward. It often requires meticulous engineering and access to a comprehensive set of task-specific information. This challenge becomes even more pronounced in naturally sparse-reward environments. In such contexts, integrating LLMs to assist RL agents has emerged as a promising direction \citep{du2023guiding}. Despite the potential of this approach, LLMs may face difficulties in understanding specific environments \citep{bommasani2021opportunities,brian2022saycan}. This limitation undermines their efficacy in assisting RL agents.

% In this paper, our goal is to enhance LLMs to better understand specific environments without relying on demanding prompt engineering or directly fine-tuning LLMs, while assisting RL agents with complex decision-making tasks in such environments. To this end, we propose a novel framework, AdaRefiner, where the LLM comprehends the environment and provides guidance to the RL agent who selects fine-grained actions to accomplish tasks. To automatically refine the LLM's comprehension of the environment, we consider adding an adjustable module to help the LLM adapt to the environment. 

In this paper, our goal is to enhance LLMs to better understand specific environments without relying on demanding prompt engineering or directly fine-tuning LLMs, while assisting RL agents with complex decision-making tasks. To this end, we propose a novel framework, AdaRefiner, where the LLM provides guidance to the RL agent who selects fine-grained actions to accomplish tasks. Simultaneously, the RL agent contributes adaptive feedback, enriching the LLM’s understanding of the environment through an adjustable module.

As illustrated in Figure \ref{fig:pipeline}, the core feature of AdaRefiner is the integration of a lightweight Adapter LM. This Adapter LM, enriched with feedback and information from the RL agent, automatically prompts a Decision LLM, like GPT-4 \citep{openai2023gpt4}. It enables a refined understanding of the environment and agents' learning capabilities without the need to alter the Decision LLM's parameters. This approach maintains the generalization abilities of LLMs while providing targeted assistance for RL agents with specific tasks. By the synergy of LLMs and RL feedback, AdaRefiner addresses the limitations of existing methods, setting a new paradigm in the integration of advanced LLMs with reinforcement learning.

%Inspired by the mentioned research and identified problems, our goal is to enhance the decision-making capabilities of LLMs through feedback but without relying on complex prompt engineering. Additionally, we aim to integrate the ability of RL to select actions at a fine-grained level. To achieve these objectives, we develop a novel framework that automatically refines its self-comprehension of tasks. This refinement is driven by feedback from RL algorithms, enabling the framework to adapt more effectively to the current RL agents being assisted.

%To overcome the aforementioned challenges of fine-tuning LLMs, our approach does not focus on directly modifying LLMs but instead considers adding adjustable modules to help LLMs adapt to tasks. This key insight led us to design AdaRefiner, a framework that enhances the synergy between LLMs and RL feedback. As illustrated in Figure \ref{fig:pipeline}, the core feature of AdaRefiner is the integration of a lightweight Adapter LM. This Adapter LM, enriched with feedback and information from RL agents, automatically prompts a Decision LLM like GPT-4. It enables a refined understanding of tasks and agents' learning capabilities without the need to alter the Decision LLM's parameters. This approach maintains the generalization abilities of LLMs while providing targeted assistance for specific tasks. By effectively bridging the gap between LLMs and RL feedback, AdaRefiner addresses the limitations of existing methods, setting a new paradigm in the integration of advanced language models with reinforcement learning feedback.

In the experiments, AdaRefiner is evaluated on $22$ tasks within the \textit{Crafter} environment \citep{crafter}. The results not only demonstrate AdaRefiner's superior performance compared to state-of-the-art baselines but also highlight its ability to guide agents towards common-sense behaviors. 

Our key contributions are summarized as follows: \textbf{1)} We propose a novel framework that aligns LLMs with downstream tasks and guides agents to effectively learn complex tasks without the need for intricate prompt engineering or intensive fine-tuning; \textbf{2)} We design the Adapter LM that correlates its own update with the learning progress of the agent and automatically generates appropriate prompts for the Decision LLM, thereby forming a feedback loop together with LLMs and RL agents; \textbf{3)} We thoroughly evaluate our framework's efficacy on $22$ diverse tasks and provide a comprehensive analysis of the experimental results.

\section{Related Work}

\textbf{Large Language Models (LLMs). }
Recent advancements in natural language processing have been significantly shaped by the emergence of LLMs. The GPT series, notably, has garnered attention for its broad task versatility, while other models like PALM and LaMDA have also contributed to the field with their unique capabilities \citep{chowdhery2022palm, thoppilan2022lamda}. A pivotal development in the evolution of LLMs is the implementation of instruction tuning \citep{instruct}, which has markedly enhanced adaptability in complex scenarios, particularly in zero-shot and few-shot learning applications. The open sourcing of some LLMs \citep{zeng2022glm, touvron2023llama} has spurred efforts in task-specific fine-tuning \citep{wu2023bloomberggpt}. While this approach often boosts task performance, it can simultaneously reduce the models' generalization abilities \citep{wang2022preserving}. Our work navigates this challenge by dynamically fine-tuning a lightweight Adapter LM via real-time feedback from RL agents, aiming to strike a balance between task-specific improvement and broad applicability. This method tailors the LLM for specific tasks while maintaining LLM's broad adaptability to new environments, addressing a key limitation in current applications.

\textbf{LLMs for RL. }
Incorporating language models to represent goals in RL utilizes the extensive knowledge of LLMs trained on large corpora. The use of LM-encoded goal descriptions has been shown to significantly improve the generalization capabilities of instruction-following agents \citep{chan2019actrce,hill2020human}. This is achieved by enabling agents to interpret and act upon complex instructions more effectively. Furthermore, pre-trained LLMs provide nuanced guidance through sub-goals and sub-policies, enhancing agent strategies and decision-making in various scenarios \citep{lynch2020language, sharma2021skill}. Subsequent research efforts have linked these sub-policies to address more intricate tasks in RL environments \citep{huang2022language,huang2022inner}. Several methods also leverage LLMs to generate intrinsic rewards, boosting the efficiency and effectiveness of RL learning \citep{choi2022lmpriors, du2023guiding}. However, the application of these methods in simple text-based games often does not transfer well to more complex and dynamic environments, leading to scalability and generalization issues \citep{zhong2021silg,wang2021grounding}. Our work addresses these challenges by making LLMs more adaptable and practical for use in sophisticated environments. The AdaRefiner framework is specifically designed to enhance the flexibility and effectiveness of LLMs, providing tailored assistance to RL agents in navigating and mastering complex decision-making tasks.

\textbf{LLMs for Open-World Games. }
Open-world games pose unique challenges, such as managing long horizons \citep{crafter} and balancing multiple objectives \citep{wang2023describe}. These complexities require sophisticated decision-making strategies. While some studies have explored using LLMs for planning and guiding RL agents \citep{du2023guiding,yuan2023plan4mc,tsai2023can}, their approaches often depend on human-generated trajectories as context. This dependency can limit the agent's performance in unseen scenarios, making them less effective compared to recent RL algorithms \citep{hafner2023mastering} that operate independently of LLMs. Additionally, methods that solely rely on LLMs for decision-making \citep{wu2023spring, wang2023voyager} often have designs that are intricately tailored to specific environments or require expert-level prior knowledge. This specificity can make them less transferable to different tasks. In contrast, our AdaRefiner avoids such complexity. Its straightforward and flexible design enables it to adapt to a variety of tasks and environments, addressing the key limitations of current LLM applications in open-world games.

\section{Methodology}

\subsection{Problem Formulation}

In our study, the primary goal is to leverage LLMs to enhance the decision-making capabilities of RL agents in complex environments. We consider a partially observable Markov decision process (POMDP), defined by the tuple $(\mathcal{S}, \mathcal{A}, \mathcal{P}, \Omega, \mathcal{O}, R, \gamma)$. Here, $s\in\mathcal{S}$ and $a\in\mathcal{A}$ denote the state and action, respectively. The transition probability $\mathcal{P}(s^\prime|s,a)$ represents the environment dynamics, where $s^\prime$ is the state following action $a$ from state $s$. The observation $o\in\Omega$ is obtained through function $\mathcal{O}(o|s,a)$, and $R$ is the reward function, with $\gamma$ as the discount factor. We can use $\tau \doteq \{o_0,a_0,r_0,\ldots,o_t, a_t, r_t,\ldots\}$ to represent a sequence of data as a trajectory.

Under this setting, we employ LLMs to generate sub-goals $g$, aiding agents in decision-making processes. These sub-goals are designed to provide intermediate targets, enhancing the agent's ability in complex scenarios. Our objective is to develop a policy, denoted as $\pi(a|o,g)$, which maximizes cumulative reward by effectively integrating these sub-goals. The specific mechanics of how LLMs assist in generating these sub-goals and their exact role in the decision-making process will be detailed in subsequent sections.

\begin{figure*}[t]
\centering
\includegraphics[width=.8\textwidth]{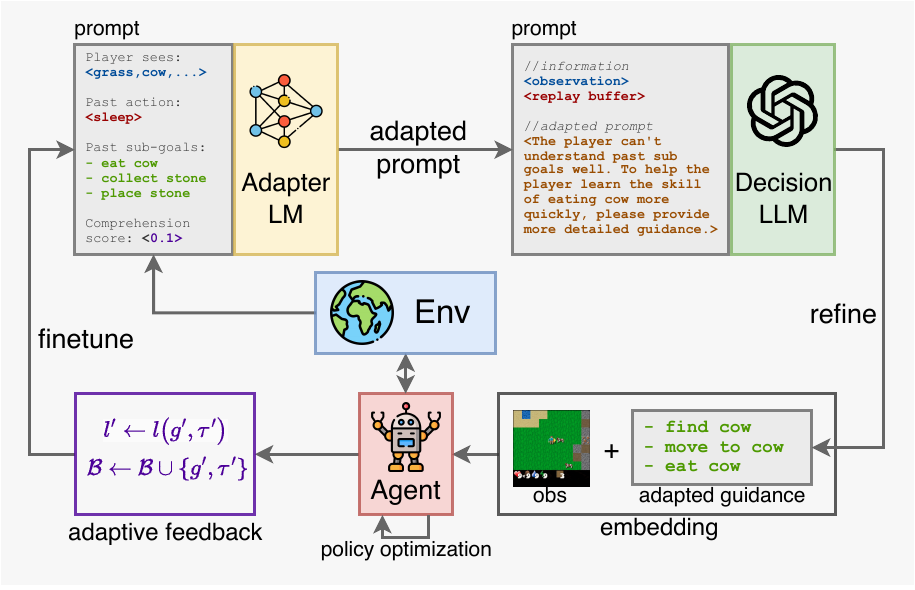}
\caption{Overall framework of AdaRefiner. In addition to receiving inputs from the environment and historical information, the prompt of the Adapter LM incorporates a comprehension score. This score computes the semantic similarity between the agent's recent actions and the sub-goals suggested by the LLM, determining whether the agent currently comprehends the LLM's guidance accurately. Through the agent's feedback and continuously fine-tuning the Adapter LM, we can keep the LLM always attuned to the actual circumstances of the task. This, in turn, ensures that the provided guidance is the most appropriate for the agents' prioritized learning.}
\label{fig:framework}
\end{figure*}

\subsection{Key Idea and Overall Framework}

Pre-trained LLMs demonstrate impressive zero-shot language understanding capabilities across diverse tasks. This proficiency can be leveraged to help agents quickly comprehend complex environments, thus mitigating exploration dilemmas in RL. By prompting LLMs, we obtain sub-goals in textual format, which are then embedded with the agent’s observations to inform the policy \(\pi(a|o,g)\). This process aids agents in making more informed decisions based on the contextual guidance provided by these sub-goals.

Despite their generalization capabilities, LLMs may not always have a comprehensive understanding of specific tasks, leading to potential mismatches between the generated guidance and the environment’s realities. Directly using LLM-generated guidance may not result in coherent or relevant advice. While fine-tuning LLMs with task-specific data is a typical solution, it can be computationally intensive and may also lead to catastrophic forgetting of pretrained knowledge. Moreover, fine-tuning black-box models like GPT-4 is infeasible due to restricted access to their weights.

Given these challenges, we focus on adding adjustable modules to help LLMs adapt to environments, rather than modifying the LLMs directly. A key insight is that even a lightweight LM, with the right fine-tuning, can excel at particular tasks \citep{zhang2023llama, li2023flexkbqa}. This motivates us to propose AdaRefiner, as illustrated in Figure \ref{fig:framework}. The core component of AdaRefiner is a lightweight Adapter LM which bridges the gap between specific environments and the Decision LLM’s capabilities. The Adapter LM first processes the environmental inputs and the agent’s status, automatically generating tailored prompts that include summaries and suggestions. These prompts are then fed into the Decision LLM, which produces final sub-goals. The Adapter LM thus acts as an intermediary, ensuring that the Decision LLM receives contextually relevant information, enabling it to provide accurate and useful guidance to the agent.

\subsection{Adapter LM}

The Adapter LM processes two types of input information: environmental information and the agent's comprehension level of language guidance. The environmental information, sourced from the game engine or visual descriptors \citep{radford2021learning}, includes critical information such as object properties and the current status of the agent. The agent’s comprehension level of language guidance is quantified using a cosine similarity score \(l\), calculated between the suggested sub-goals and the agent's trajectories, represented as:
\begin{equation}
l\doteq\cos(g,\tau)=\frac{f_\text{emb}(g)\cdot f_\text{emb}(\tau)}{\|f_\text{emb}(g)\|\cdot \|f_\text{emb}(\tau)\|}.
\end{equation}
Here, \(f_\text{emb}\) represents the embedding function, with SentenceBert \citep{reimers-2019-sentence-bert} employed in our implementation. A higher score \(l\) suggests that the agent’s actions are more closely aligned with the sub-goals, indicating a better comprehension of the provided guidance.

The Adapter LM then utilizes the comprehension score \(l\) and environmental information to generate \(\texttt{prompt}_\texttt{a}(\mathcal{B},l)\), where \(\mathcal{B}\) is a replay buffer of the agent’s historical contexts and \(\texttt{prompt}_\texttt{a}(\cdot)\) is the prompt template for Adapter LM. After analyzing the prompt, the Adapter LM synthesizes the information to assist the Decision LLM, which is responsible for overall decision-making. The output from the Adapter LM, represented as \(c{\sim}\mathcal{M}_\text{a}(\texttt{prompt}_\texttt{a}(\mathcal{B},l))\), is then used to inform the Decision LLM. Here, \(\mathcal{M}_\text{a}\) represents the Adapter LM. By providing tailored information through the adapted \(\texttt{prompt}_\texttt{d}(\mathcal{B},c)\), the Decision LLM is better equipped to generate appropriate sub-goals \(g{\sim}\mathcal{M}_\text{d}(\texttt{prompt}_\texttt{d}(\mathcal{B},c))\). Here, \(\mathcal{M}_\text{d}\) represents the Decision LLM and \(\texttt{prompt}_\texttt{d}(\cdot)\) is the prompt template for the Decision LLM. Details of these prompts is available in Appendix \ref{ap:prompts}.

\subsection{Training Procedure}

The training process of our framework is designed to coordinate the learning of RL agents and the fine-tuning of the Adapter LM. In other words, the Adapter LM is continuously updated to refine its comprehension of the environment and the agent in parallel with the RL agent's exploration and data collection. Specifically, the RL agent receives suggested sub-goals \(g{\sim}\mathcal{M}_\text{d}(\texttt{prompt}_\texttt{d}(\mathcal{B},c))\) from the Decision LLM, which are then provided to the policy \(\pi(a|o,g_\text{emb})\) for training. Here, \(g_\text{emb}\) is the text embedding produced by \(f_\text{emb}\). The agent’s actions and the resultant trajectories provide an updated comprehension score \(l^\prime = \cos(g^\prime,\tau^\prime)\), where \(g^\prime,\tau^\prime\) is the new sub-goals and trajectories. This score and collected information are then used to compose a linguistic data pair \(\text{<}~\texttt{prompt}_\texttt{a}(\mathcal{B},l^\prime), c ~\text{>}\) for supervised fine-tuning of the Adapter LM. Then the replay buffer will be updated as \(\mathcal{B}\leftarrow\mathcal{B}\cup\{g^\prime,\tau^\prime\}\) This iterative procedure allows the Adapter LM to continuously refine its self-awareness and generate more effective summaries \(c\), which affects the quality of guidance for the RL agent.

Considering the computational costs and the nature of open-world game environments, we query the language models at predetermined intervals instead of every step. This strategy ensures a balance between consistent guidance and computational efficiency. The fine-tuning of the Adapter LM is also conducted at specific intervals for the same reason. In line with our claim that only a lightweight Adapter LM is needed, we utilize the 4-bit quantized version of the Llama2-7B model \citep{touvron2023llama2} as the base model \citep{jiang2023mistral} and employ QLoRA \citep{dettmers2023qlora} for efficient fine-tuning. And we choose OpenAI's GPT-4 as the default Decision LLM. These choices will be further discussed and analyzed in Section \ref{sec:exp}. For policy learning, we adopt the classic Proximal Policy Optimization (PPO) algorithm \citep{schulman2017proximal}. It is worth noting that our framework is designed to be compatible with a variety of standard RL algorithms, not only limited to PPO. 

Specific parameters and settings are detailed in Appendix \ref{ap:details}. The complete procedure can be found in Appendix \ref{sec: pseudo}.

\section{Experiment}\label{sec:exp}

Our experiments primarily aim to validate the following claims: \textbf{1)} The integration of the Adapter LM can enhance LLM's comprehension of downstream tasks and the agent's understanding capability, resulting in more meaningful guidance; \textbf{2)} Agents trained under the AdaRefiner framework can exhibit superior performance and demonstrate higher-level decision-making capabilities.

\subsection{Experiment Settings}

Our experiments are conducted in the \textit{Crafter} environment \citep{crafter}, a widely used benchmark with $22$ different tasks for evaluating the decision-making capabilities of agents in open-world games. 

\textbf{Environment Details.} 
\textit{Crafter} features a $64\times 64$ grid map populated with various objects (\eg, grass, water, wood) and entities (\eg, player, zombie, skeleton). Agents in this environment have access to a local $9\times 7$ area for observation, presenting a challenge in terms of limited information and requiring effective decision-making for long-term survival and resource management. In \textit{Crafter}, agents are not bound to a single main task. Instead, they are expected to master a range of skills to accomplish $22$ different tasks, including tasks such as collecting resources, crafting tools, and surviving against environmental hazards. This variety tests the agents' ability to learn and adapt to diverse challenges, aligning well with our objective to enhance their decision-making capabilities through the AdaRefiner framework.

\textbf{Evaluation Metrics.}
In \textit{Crafter}, the performance of an agent is evaluated using three metrics: reward, success rate, and overall score. The reward is designed to reflect the agent's skills. Each time an agent unlocks a new achievement, it receives a $+1$ reward. Additionally, the agent is rewarded with $+0.1$ or penalized with $-0.1$ for every gain or loss of a health point, respectively. The success rate is defined as the proportion of episodes in which agents complete a achievement. Completing the same achievement multiple times within an episode does not affect the success rate. The overall score averages the success rates ($s_i\in[0,100]$) of the $22$ achievements in log-space as follows (known as the geometric mean): $S\doteq\exp{\left(\frac{1}{N}\sum_{i=1}^{N}\ln{\left(1+s_i\right)}\right)}-1$, where $N=22$ is the total number of achievements.
% This score metric provides a measure of the agent's overall performance, encapsulating its ability to handle a variety of tasks and challenges within the environment.

% \textbf{LLM Choices.}
% In the AdaRefiner framework, we have selected OpenAI's GPT-4 as the default Decision LLM due to its advanced natural language processing capabilities and demonstrated effectiveness in a wide range of AI applications. In our ablation study, we will explore the specific impacts that different choices of LLMs have on providing decision-making assistance to agents.

\textbf{Prompt Design.}
The prompt design for the Adapter LM is crafted to encapsulate critical information for decision-making. It includes observations of objects and the agent's status obtained from the game engine, along with the comprehension score \(l\). The format is: ``\texttt{Player sees: <observations>;~Player status: <status>;~Past action: <past actions>;~Past sub-goals: <last suggested sub-goals>;~Comprehension score: <\(l\)>. Analyze the environment and the player's understanding capability, then generate concise summaries and suggestions about this player.}'' For the Decision LLM, we construct the prompt based on the Adapter LM's output: ``\texttt{<output of the Adapter LM>. Based on the provided information, suggest 3 sub-goals that the player should accomplish next.}''

\subsection{Baselines}

To demonstrate the effectiveness of our AdaRefiner framework, we conduct comparative analyses against a diverse set of methods:

\textbf{LLM-based Methods}: We compare AdaRefiner with LLM-based methods such as Reflexion \citep{shinn2023reflexion}, ReAct \citep{yao2023react}, and Vanilla GPT-4. Reflexion and ReAct leverage chain-of-thought prompts for decision-making tasks. Considering that LLM-based methods do not accept image input, we additionally include the coordinates of objects in the prompt for fair comparisons. 
We also maintain consistency in the prompts used across all methods (detailed in Appendix \ref{ap:prompts}).
The LLM used in Reflexion and ReAct is the same as the Decision LLM in AdaRefiner, \ie, GPT-4 by default.
These comparisons aim to demonstrate how the integration of LLMs with adaptive feedback can provide a more comprehensive approach to decision-making.

\textbf{RL Methods}: We also benchmark against RL methods such as DreamerV3 \citep{hafner2023mastering}, Rainbow \citep{hessel2018rainbow}, PPO \citep{schulman2017proximal}, RND \citep{burda2019exploration}, and Plan2Explore \citep{sekar2020planning}. DreamerV3 is notable for its performance in model-based RL. Rainbow is a classic algorithm that achieves great performance in many games. RND and Plan2Explore are known for the intrinsically motivated exploration. PPO, which is also adopted in AdaRefiner, serves to highlight the added value of LLMs in the same RL setup.

%\textbf{Additional References}: We include human expert performance \citep{crafter} and standalone GPT-4 (providing step-by-step instructions) for additional references. We also include a random policy as a basic reference to contextualize the effectiveness of all the methods in the experiment.

\textbf{Additional References}: We include random policy, human expert performance \citep{crafter}, SPRING \citep{wu2023spring} that provides GPT-4 with domain-specific prior knowledge (\ie, research papers about the game engine), and Reflexion with \textit{gpt-4-vision} (including both coordinates and image inputs), as additional references to showcase performances enhanced with different information and knowledge.

% Comparing AdaRefiner with these baselines can appropriately validate the efficacy of AdaRefiner in various aspects, such as decision-making capability and overall performance in the complex \textit{Crafter} environment.

\begin{table}[t]
\resizebox{\columnwidth}{!}{
\setlength{\tabcolsep}{4pt}
\begin{tabular}{clcc}
\toprule[1pt]
\textbf{Method Type} &\textbf{Method}          & \textbf{Score (\%)}             & \textbf{Reward} \\
\midrule
\multirow{2}{*}{Ours}& AdaRefiner (@5M)   & $\boldsymbol{28.2}\pm1.8$    & $\boldsymbol{12.9}\pm 1.2$ \\
& AdaRefiner (@1M)       & $ 15.8 \pm 1.4$ & $12.3\pm 1.3$ \\
\midrule
\multirow{3}{*}{\shortstack{LLM-based\\ methods}} & Reflexion (GPT-4)       & $11.7\pm 1.4$   & $9.1\pm 0.8$  \\
& ReAct (GPT-4)          & $8.3\pm 1.2$    & $7.4\pm 0.9$  \\
% & ELLM            & N/A               & $6.0 \pm 0.4$ \\
& Vanilla GPT-4           & $3.4 \pm 1.5$   & $2.5\pm 1.6$   \\
\midrule
\multirow{5}{*}{\shortstack{RL methods}} & DreamerV3       & $14.5 \pm 1.6$  & $11.7 \pm 1.9$ \\
& PPO             & $4.6 \pm 0.3$   & $4.2 \pm 1.2$  \\
& Rainbow         & $4.3 \pm 0.2$   & $5.0 \pm 1.3$  \\
& Plan2Explore    & $2.1 \pm 0.1$   & $2.1 \pm 1.5$  \\
& RND             & $2.0 \pm 0.1$   & $0.7 \pm 1.3$  \\
\midrule
\midrule
\multirow{4}{*}{\shortstack{Addtional\\ references}} & Human Experts   & $50.5 \pm 6.8$  & $14.3 \pm 2.3$ \\
& SPRING (+prior) & $27.3\pm1.2$    & $12.3\pm 0.7$ \\
& Reflexion (GPT-4-Vision) & $12.8 \pm 1.0$ & $10.3 \pm 1.3$ \\
& Random          & $1.6 \pm 0.0$   & $2.1 \pm 1.3$   \\
\bottomrule[1pt]
\end{tabular}
}
\caption{Performance comparison between AdaRefiner and baselines in terms of score and reward metrics. AdaRefiner is trained with $5$ million and $1$ million steps. All results are derived from $500$ inference episodes. Note that $\pm$ captures standard deviations.} %The low success rates for these two tasks may be due to the numerous prerequisite conditions required for level-$7$ tasks, resulting in their less frequent representation in the statistics.}
\label{tab:perf}
\end{table}

\subsection{Results and Analysis}\label{sec:results}

The comparison includes some methods for which open-source codes are unavailable. For these algorithms, we rely on the performance metrics reported in respective papers, ensuring that the comparisons are as consistent as possible in terms of experimental setup and evaluation criteria. 
For RL baselines, we set the training to $1$ million steps, following the standard set in the \textit{Crafter} paper \citep{crafter}. However, LLM-based baselines do not include a training phase and instead focus on leveraging pre-trained LLMs. To facilitate a fair comparison, we also present a version of AdaRefiner trained for $5$ million steps to assess its asymptotic performance. This extended training is essential for evaluating AdaRefiner's full potential and maintaining comparability with baselines.

Results in Table \ref{tab:perf} show that AdaRefiner with $1$ million training steps outperforms all baselines. In comparisons with RL methods, the integration of LLM demonstrates a clear advantage in learning effectiveness. The performance of AdaRefiner compared to Reflexion and ReAct underscores that prompts generated automatically by the Adapter LM can enhance the decision-making capabilities of LLMs in downstream tasks more effectively than traditional prompt engineering techniques. This efficiency, combined with the adaptability of AdaRefiner, establishes it as a highly practical and powerful framework in complex decision-making environments.

% Although the baseline SPRING, which \textit{\textbf{requires inputting task-related research papers and 9-round chain-of-thought questioning}}, shows performance slightly lower than AdaRefiner, its reliance on expert-level prior knowledge and intensive computational resources makes it less practical. In contrast, AdaRefiner achieves top performance through a more streamlined approach. It utilizes the Adapter LM to automate a two-step prompting process, significantly reducing the complexities. 

\begin{table}[t]
\begin{center}
\resizebox{.9\columnwidth}{!}{
\setlength{\tabcolsep}{4pt}
\begin{tabular}{lcc}
\toprule[1pt]
\textbf{Method}          & \makecell{\textbf{Achievements}\\(out of $22$)} & \makecell{\textbf{Achievement Depth}\\(max $8$)} \\
\midrule
AdaRefiner       & \textbf{21}           & \textbf{7}                \\
DreamerV3       &      19      &         6        \\
Reflexion &     17       &         5        \\
\bottomrule[1pt]
\end{tabular}
}
\caption{Numbers and depths of achievements that can be completed by different methods. The achievement depth refers to the number of prerequisite steps required to complete each task, with a maximum value of $8$.}% and can indicate the difficulty of each task.}
\label{tab:num-achievements}
\end{center}
\end{table}

\begin{figure*}[t]
\centering
\includegraphics[width=\textwidth]{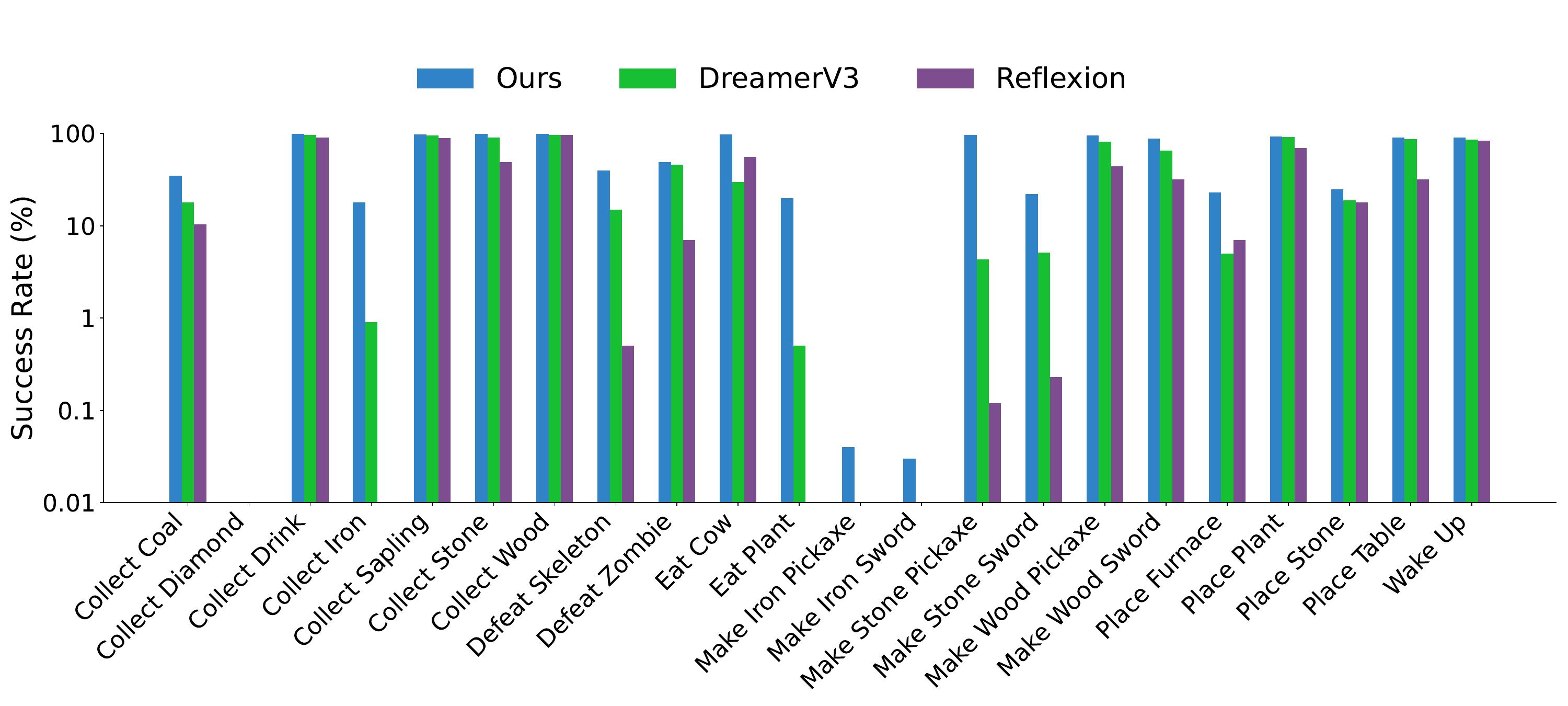}
\caption{Success rates of unlocking $22$ different achievements in log scale. AdaRefiner outperforms the two top-performing baselines. Notably, AdaRefiner is the only method that successfully completes the level-$7$ tasks ``Make Iron Pickaxe'' and ``Make Iron Sword''.}
\label{fig:success-rate}
\end{figure*}

Additionally, it is noteworthy that AdaRefiner with $5$ million training steps slightly outperforms the SPRING (+prior) method, which necessitates providing task-related papers and engaging in a 9-round chain-of-thought questioning process. This indicates that AdaRefiner can achieve better performance simply through a comprehensive understanding of adaptive feedback, without the need for external expert-level knowledge and complex prompt engineering.

Moreover, the augmented Reflexion, which utilizes \textit{gpt-4-vision}, showes some improvements over the original version. However, even with the inclusion of additional input information, Reflexion still exhibited a significant performance gap compared to AdaRefiner. This indicates that merely employing pretrained large multimodal models with image understanding capabilities does not guarantee improved performance on downstream tasks, further underscoring the efficacy of AdaRefiner's framework design.

% In the investigation of success rates on $22$ specific achievements within \textit{Crafter}, AdaRefiner and two top-performing baselines, DreamerV3 and Reflexion. As depicted in Figure \ref{fig:success-rate}, AdaRefiner shows higher success rates in most achievements. Exceptions are ``Defeat Skeleton'' and ``Defeat Zombie'', where its success rates are slightly lower. Replay analysis reveals that AdaRefiner's policy tends to prioritize survival over combat. This strategy, emerging from the collaboration between the Adapter LM and GPT-4, illustrates an unexpected but practical approach to the game's challenges.

To study the breadth of abilities learned by different methods, we compare AdaRefiner with two top-performing baselines, DreamerV3 and Reflexion. We investigate their success rates on $22$ specific achievements in \textit{Crafter}. Both AdaRefiner and DreamerV3 are trained for $5$ million steps. Figure \ref{fig:success-rate} illustrates that AdaRefiner has the highest success rates across all tasks. Moreover, as shown in Table \ref{tab:num-achievements}, AdaRefiner completes the largest number of achievements and is the only method that reaches level-$7$ difficulty. Specifically, AdaRefiner is notably the only method capable of accomplishing level-$7$ tasks ``Make Iron Pickaxe'' and ``Make Iron Sword''. These tasks are particularly hard due to their prerequisite conditions and rarity in the game. This result underscores the importance of a comprehensive understanding of environments in developing versatile agents.

\subsection{Ablation Study}

% \begin{table}[ht]
% \caption{Ablation study results of AdaRefiner. Achievement depth represents the highest level of achievement that agents can accomplish, with a maximum value of $7$. These methods indicate the impact on performance caused by using different LLMs and whether or not to include the Adapter LM. All results are obtained from $5$ independent trials.}
% \label{tab:ablation}
% \resizebox{\columnwidth}{!}{
% \begin{tabular}{lccc}
% \toprule[1pt]
% \textbf{Method}                   & \textbf{Score}                       & \textbf{Reward}                     & \textbf{Achievment Depth} \\
% \midrule
% AdaRefiner                & $\boldsymbol{28.2\pm1.8\%}$ & $\boldsymbol{12.9\pm 1.2}$ & 7                \\
% AdaRefiner w/ GPT-3.5     & $ 25.8\pm 2.0\%$            & $ 12.5\pm 1.5$             & 6                \\
% \midrule
% AdaRefiner                & $\boldsymbol{28.2\pm1.8\%}$ & $\boldsymbol{12.9\pm 1.2}$ & 7                \\
% AdaRefiner w/o u-score    & $ 13.4\pm 1.9\%$            & $ 9.2\pm 1.6$              & 5                \\
% AdaRefiner w/o Adapter LM & $ 9.6\pm 1.7\%$             & $ 8.7\pm 1.4$              & 5                \\
% \midrule
% AdaRefiner                & $\boldsymbol{28.2\pm1.8\%}$ & $\boldsymbol{12.9\pm 1.2}$ & 7                \\
% GPT-4 + GPT-4            & $7.5\pm 0.8\%$                & $ 5.2\pm 1.5$              & 4                \\
% Llama2-7b + GPT-4 & & & \\
% \bottomrule[1pt]
% \end{tabular}
% }
% \end{table}

To investigate the contribution of various components in the AdaRefiner framework, a series of ablation studies are conducted.

\textbf{Decision LLM Variants.} We first investigate the performance of using different Decision LLMs. By replacing GPT-4 with GPT-3.5 in the Decision LLM, we observe a slight decrease in performance, as shown in the first two rows of Table \ref{tab:ablation}. This result suggests that AdaRefiner using an LLM with less capability still maintains a comparable level to other baselines, achieving level-$6$ tasks. This demonstrates that the success of AdaRefiner is primarily attributed to its framework design, rather than the use of more advanced GPT-4 as the Decision LLM. In contrast, when comparing the two versions of Reflexion under the same Decision LLMs, significant performance gaps are observed, further underscoring the superiority of our framework.

\begin{table}[t]
\resizebox{\columnwidth}{!}{
\setlength{\tabcolsep}{3pt}
\begin{tabular}{lccc}
\toprule[1pt]
\textbf{Method (@5M steps)}                   & \textbf{Score (\%)}                       & \textbf{Reward}                     & \makecell{\textbf{Achievement }\\\textbf{Depth}} \\
\midrule
AdaRefiner                & $\boldsymbol{28.2}\pm1.8$ & $\boldsymbol{12.9}\pm 1.2$ & 7                \\
\midrule
\midrule
AdaRefiner (GPT-3.5)     & $ 23.4\pm 2.2$            & $ 11.8\pm 1.7$             & 6                \\
Reflexion (GPT-4)       & $11.7\pm 1.4$   & $9.1\pm 0.8$ & 5 \\
Reflexion (GPT-3.5) & $8.9\pm 1.7$ & $7.2\pm 1.1$ & 4 \\
\midrule
% AdaRefiner                & $\boldsymbol{28.2}\pm1.8$ & $\boldsymbol{12.9}\pm 1.2$ & 7                \\
AdaRefiner w/o \(l\)-score    & $ 13.4\pm 1.9$            & $ 9.2\pm 1.6$              & 5                \\
AdaRefiner w/o Adapter LM & $ 9.6\pm 1.7$             & $ 8.7\pm 1.4$              & 5                \\
\midrule
% AdaRefiner                & $\boldsymbol{28.2}\pm1.8$ & $\boldsymbol{12.9}\pm 1.2$ & 7                \\
GPT-4 + GPT-4            & $7.5\pm 0.8$                & $ 5.2\pm 1.5$              & 4                \\
Llama2-7B + GPT-4        & $7.1 \pm 1.0$ & $4.7\pm 1.5$ & 4 \\
\midrule
% AdaRefiner                & $\boldsymbol{28.2}\pm1.8$ & $\boldsymbol{12.9}\pm 1.2$ & 7                \\
AdaRefiner w/ binary score &  $18.7 \pm 2.4$ & $11.0 \pm 1.6$ & 6 \\
\bottomrule[1pt]
\end{tabular}
}
\caption{\label{tab:ablation}Ablation study of AdaRefiner. The results illustrate the impact of various components.}
\end{table}

\begin{figure*}[t]
\centering
\resizebox{\linewidth}{!}{
\begin{tabular}[t]{@{}c@{}c@{}c}
\subfloat{\includegraphics[width=0.13\textwidth]{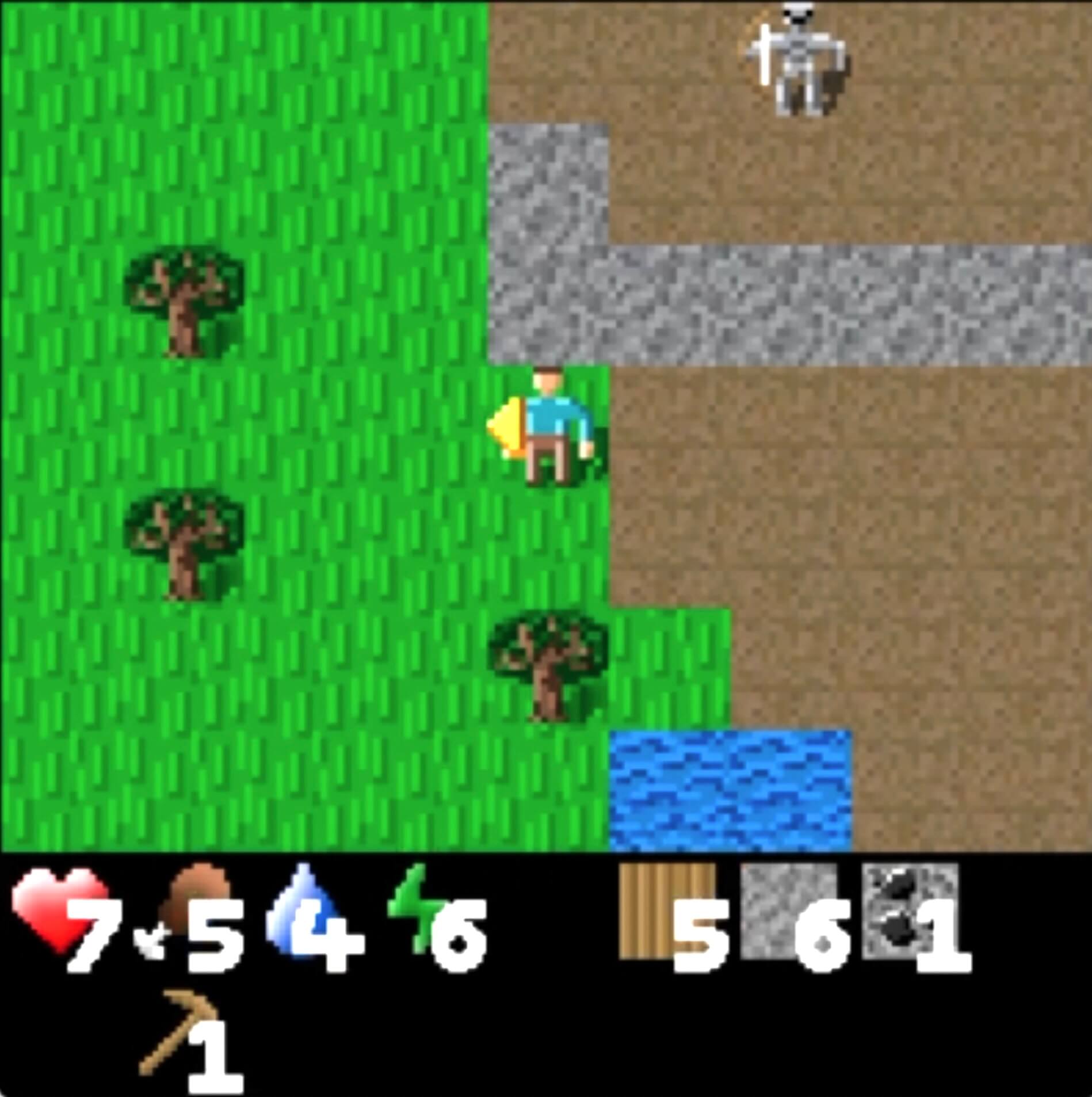}} & 
\subfloat{\includegraphics[width=0.13\textwidth]{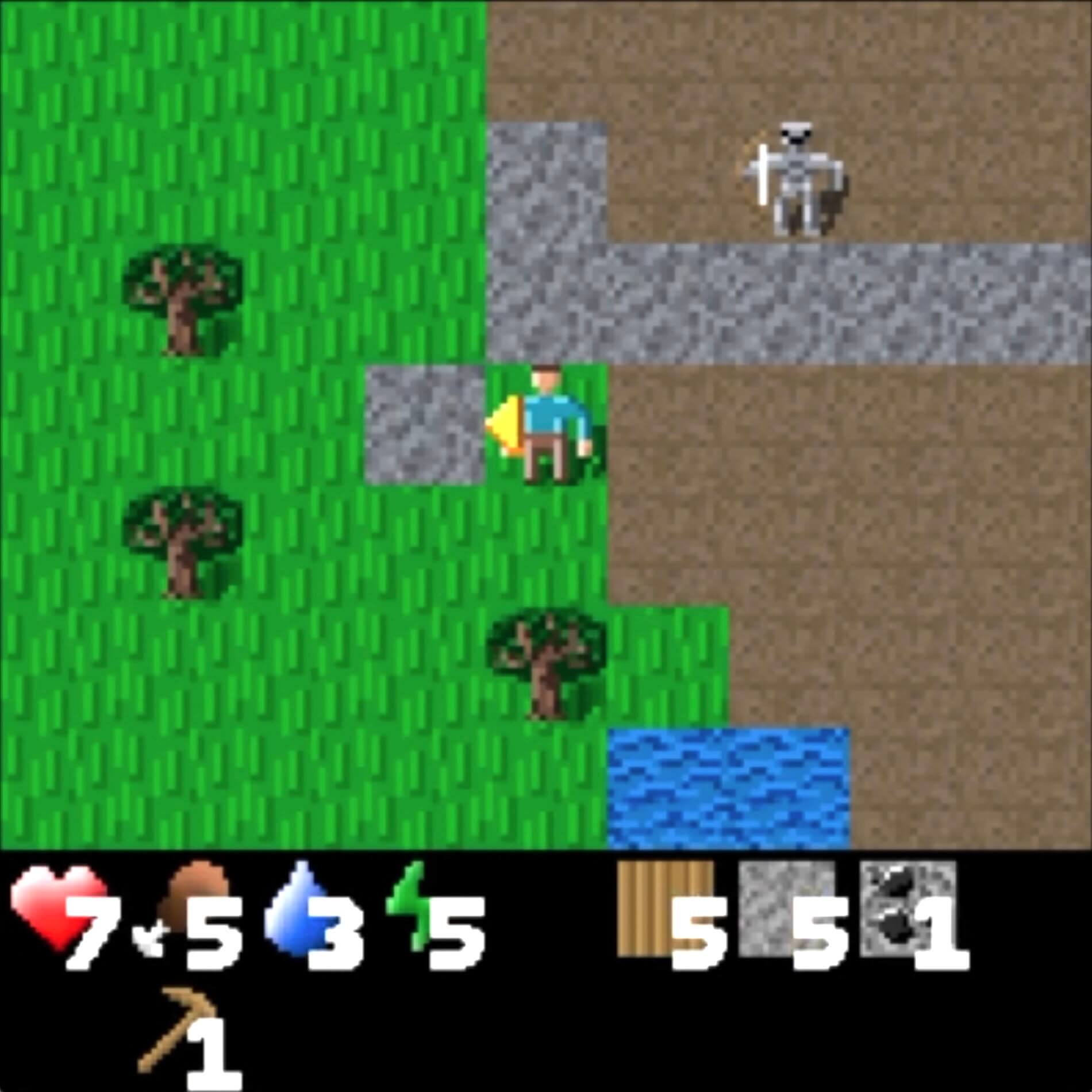}} & 
\multirow{2}{*}[5.8em]{\subfloat{\includegraphics[width=0.74\textwidth]{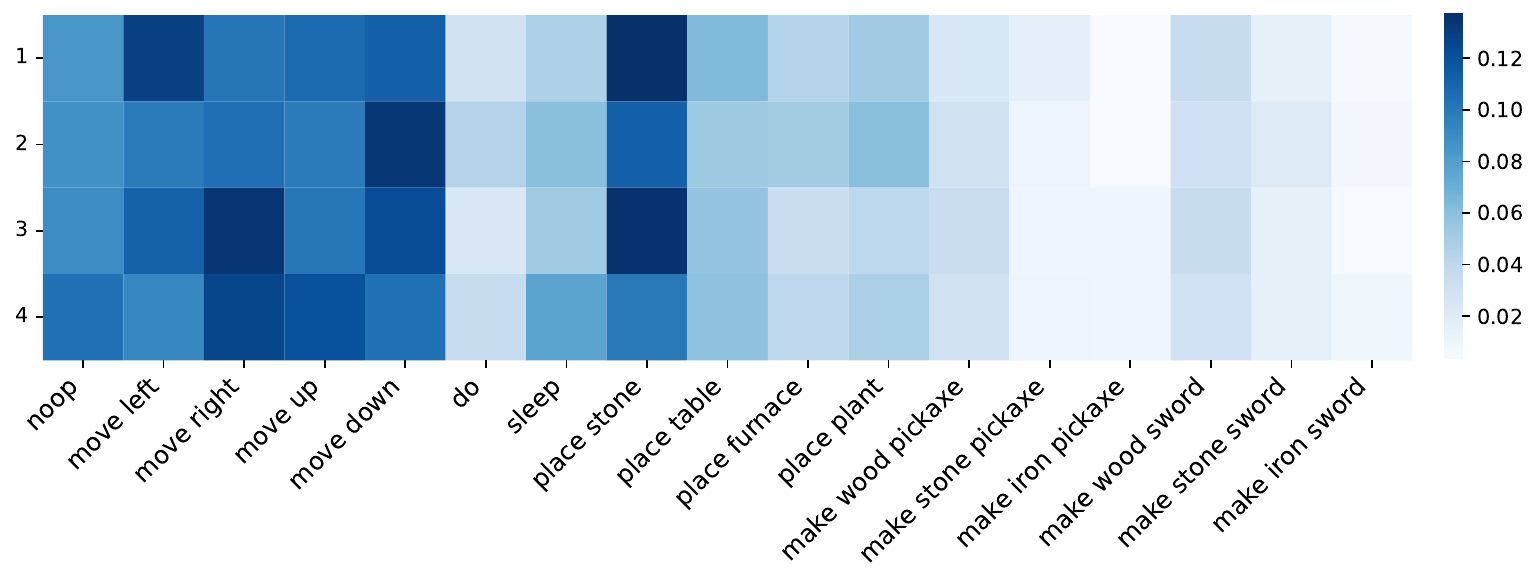}}} \vspace{-0.5cm}\\ 
\subfloat{\includegraphics[width=0.13\textwidth]{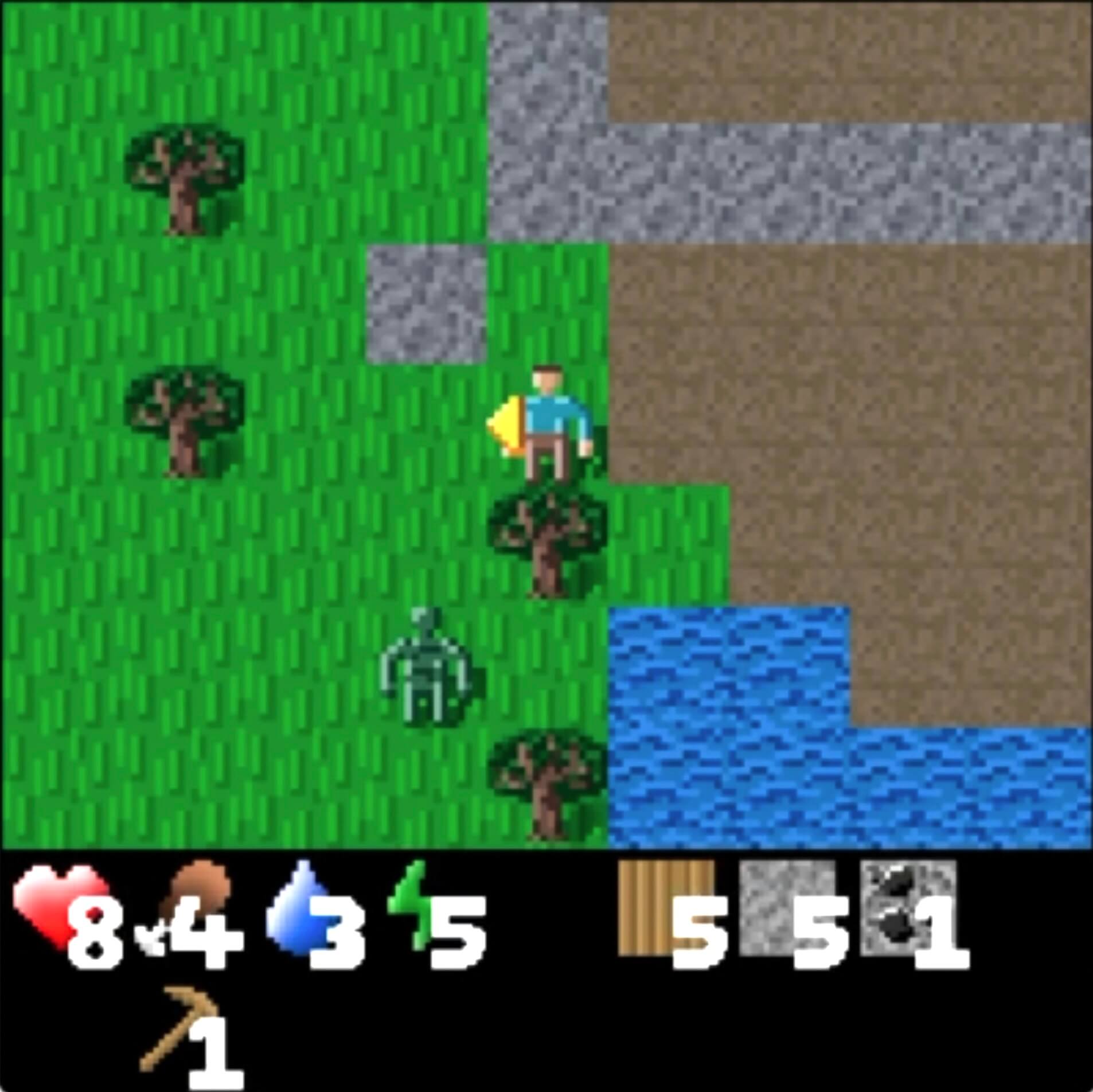}} & 
\subfloat{\includegraphics[width=0.13\textwidth]{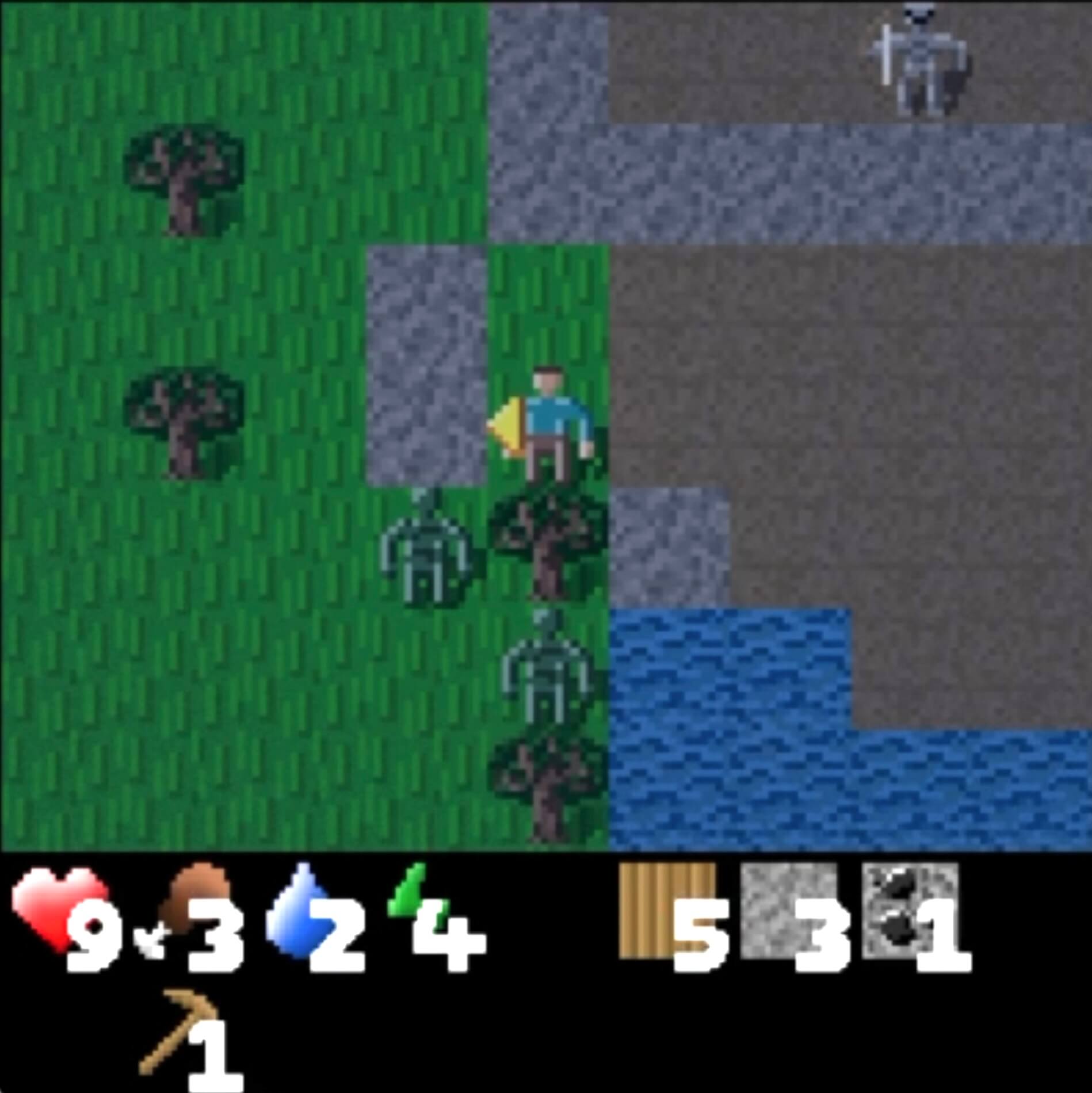}}
\end{tabular}
}
\caption{(\textit{left}) Frames from an episode in the game, the order is from top left to bottom right. (\textit{right}) The probabilities of actions in the agent's policy corresponding to each frame.}
\label{fig:case1}
\end{figure*}

\textbf{Adapter LM Variants.} To study the contribution of the Adapter LM to AdaRefiner, we design two variants as shown in the middle three rows of Table \ref{tab:ablation}. The first variant, \textit{AdaRefiner w/o l-score}, excludes the comprehension score from both the prompts and the fine-tuning process. This variant experiences a notable performance decline, highlighting the critical role of the comprehension score in refining the Adapter LM with task objectives. It appears that merely using task data for fine-tuning does not sufficiently enhance decision-making capabilities. Another variant, \textit{AdaRefiner w/o Adapter LM}, retains the comprehension score but removes the Adapter LM. This setup leads to an even more pronounced decrease in performance, indicating that simply providing comprehension scores as inputs is not enough to significantly increase decision-making effectiveness. It demonstrates that the Adapter LM, when fine-tuned with comprehension scores, plays a pivotal role in enhancing the overall decision-making capabilities.

\textbf{Feedback from RL.} To demonstrate the significance of integrating adaptive feedback from RL, we compare two variants that remove adaptive feedback from RL and rely solely on the Decision LLM for action decisions. In these variants, PPO and corresponding feedback are removed, and the Adapter LM is used only for inference, without any fine-tuning. The results are shown in the last three rows of Table \ref{tab:ablation}. The first variant, named \textit{Llama2-7B + GPT-4}, shows a significant decrease in performance. This underscores the critical role of incorporating adaptive feedback from RL for the Adapter LM to accurately perceive and adapt to the environment. Another variant, \textit{GPT-4 + GPT-4}, which utilizes GPT-4 as the Adapter LM for inference, exhibits similar performance, further suggesting that simply increasing the capacity of LLMs is insufficient. These comparisons demonstrate that the synergy between LLMs and RL feedback is crucial to the efficacy of AdaRefiner.

\textbf{Fine-grained Comprehension Score.} To verify the necessity of using a fine-grained comprehension score, we investigate the impact of the score's format on performance. Specifically, we compare the performance with a variant, \textit{AdaRefiner w/ binary score}, which assigns a score of $1$ to entries above a $0.5$ similarity threshold and $0$ to others. The results clearly show that replacing the comprehension score with a binary score leads to a significant decrease in performance. This indicates that a finer-grained similarity score is more effective in aiding the Adapter LM to understand the agent's capabilities, showcasing the Adapter LM's sensitivity to score values.

\subsection{Guidance and Agent Behaviors}\label{sec:case}

We further investigate how AdaRefiner enhances the agent's comprehension and learning. As shown on the left side of Figure \ref{fig:case1}, in a scenario where enemies gradually appear, AdaRefiner receives environmental information and suggests the agent to ``place stone to build shelter, collect food and drink, avoid combat''.  The policy visualized on the right side of Figure \ref{fig:case1}, reveals a high probability of ``place stone'' following this guidance. Notably, five basic actions controlling the player's movement also maintain high probabilities. This pattern likely reflects the inherent design of RL algorithms to encourage exploration, leading agents to consistently engage in common and easily executed actions. Actions less relevant to the provided guidance exhibit lower probabilities, indicating the agent's ability to prioritize actions based on AdaRefiner's suggestions. For more detailed analyses, statistical tests, and further demonstrations, please refer to Appendix \ref{ap:behaviors}.

% Detailed and visualized case studies are available in Appendix \ref{ap:behaviors}. These studies involve in-depth analyses of the relationship between the refinements offered by AdaRefiner and the agent's action probabilities. The findings underscore the crucial role of AdaRefiner in assisting agents to more effectively understand tasks and make informed decisions.

\subsection{Consistent Increment of Performance and Agent's Comprehension.}

\begin{figure}[t]
    \centering
    \subfloat{
        \includegraphics[width=0.48\columnwidth]{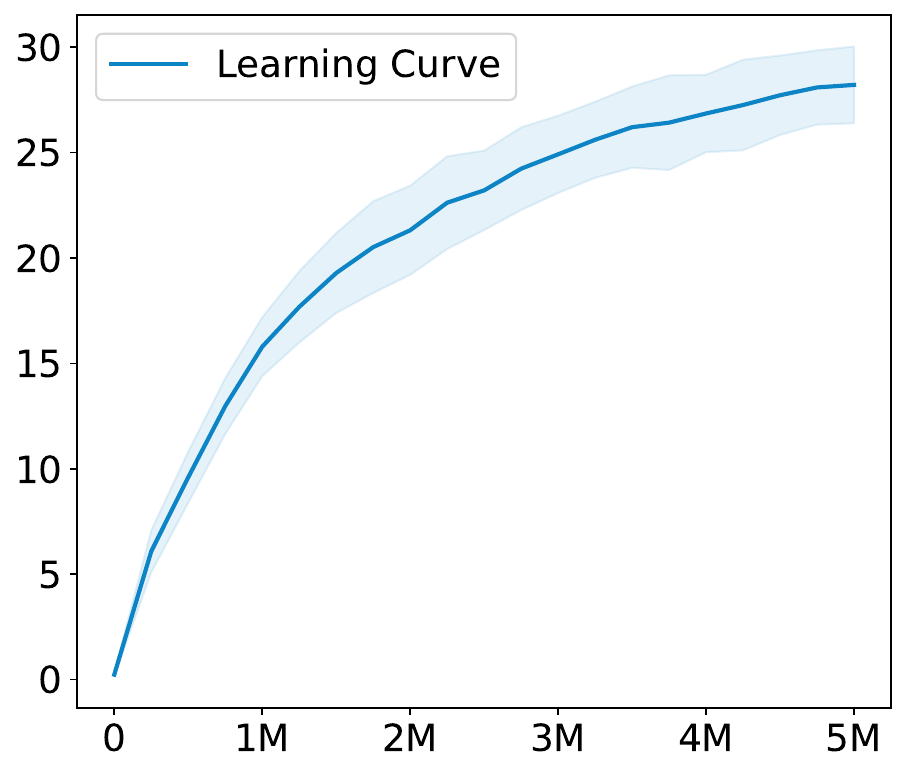}
    }
    \subfloat{
        \includegraphics[width=0.48\columnwidth]{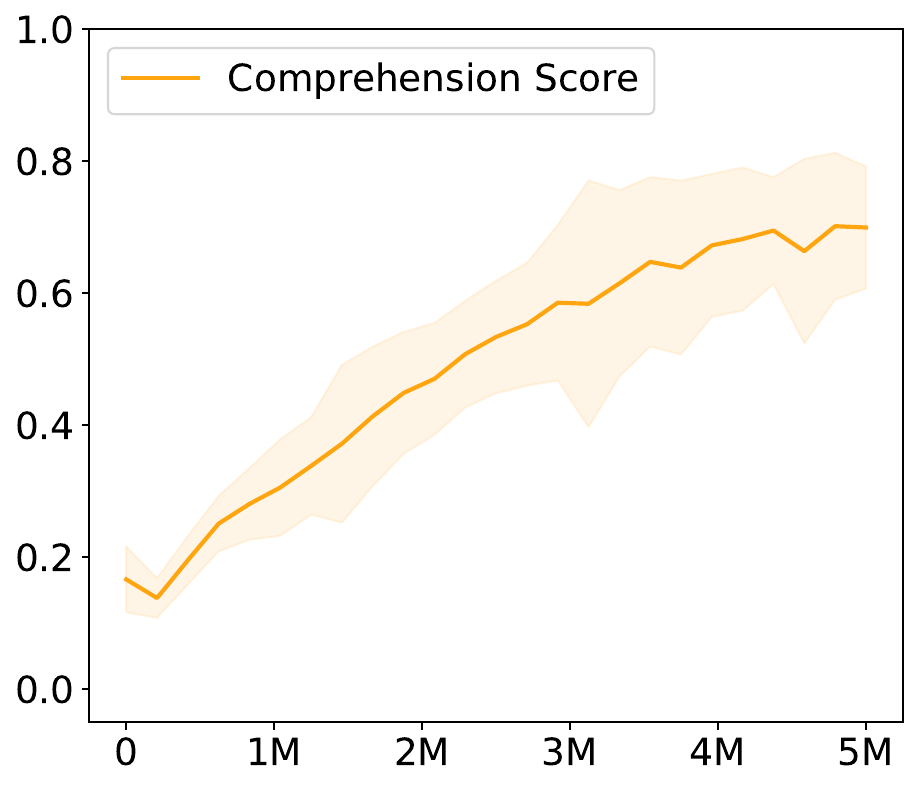}
    }
    \caption{Learning curve (\textit{left}) and comprehension score (\textit{right}) of AdaRefiner.}
    \label{fig:lc-and-cs}
\end{figure}

To further validate the efficacy of AdaRefiner in providing effective guidance for the agent, we investigate the correlation between the learning curve and the comprehension score during training. Figure \ref{fig:lc-and-cs} illustrates this relationship, showing that there is a consistent increase in the comprehension score as training progresses. This increment suggests an improvement in the agent's understanding of the language guidance, which in turn enhances the overall performance. The results demonstrate that the agent is not just following instructions more accurately but is also integrating this guidance more effectively into its decision-making process.

\subsection{Behavior Statistics}

To better quantify the guidance provided by AdaRefiner and the common-sense behavior exhibited by the agent, we have adopted a setting similar to that used in existing work \citep{du2023guiding}. Specifically, we categorized each instruction and actual agent action into three groups: 

\begin{itemize}
    \item \textbf{No Common-Sense} (where behavior significantly deviates from typical human common sense, \ie, \textit{suggesting the agent to fight with enemies when its health is low});
    \item \textbf{Impossible} (where the resources and conditions do not support the behavior in game engine);
    \item \textbf{Reasonable} (all remaining behaviors not included in the first two categories). 
\end{itemize}

The results are shown in Table \ref{tab:behavior-tests}, suggesting that in most scenarios (83.8\% and 78.6\%), the guidance and agent's actions are reasonable. While we acknowledge that there may be subjectivity in this assessment, we believe the results can still show the general tendencies of AdaRefiner in guiding agent's behavior.

\begin{table}[ht]
\centering
\begin{tabular}{ccc}
\toprule
& \textbf{Guidance} & \textbf{Action} \\
\midrule
No Common-Sense & 12.8\% & 21.4\% \\
Impossible & 3.4\% & 0\% \\
Reasonable & 83.8\% & 78.6\% \\
\bottomrule
\end{tabular}
\caption{Statistical tests on common-sense behaviors. The results are based on 500 inference samples.}
\label{tab:behavior-tests}
\end{table}

\section{Conclusions}

In this study, we introduce AdaRefiner, a novel framework that synergizes LLMs with adaptive feedback, leveraging an Adapter LM as a crucial intermediary. AdaRefiner, rigorously tested across $22$ diverse tasks in the \textit{Crafter} environment, not only outperforms state-of-the-art baselines but also steers agents towards learning higher-level skills and exhibiting common-sense behaviors. Ablation studies further validate the significance of each component, particularly emphasizing the Adapter LM's role in refining decision-making. These results highlight AdaRefiner's potential in advancing LLMs' capabilities in complex open-world games, and open up avenues for further research in LLM's decision-making capabilities.

\clearpage

\section*{Limitations}

The primary limitation of AdaRefiner is that it still requires a certain level of pre-trained knowledge of the Adapter LM. If a smaller language model is used as the Adapter LM, its language understanding ability may not be sufficient to provide the necessary analysis and summarization for the environment and agent. Additionally, although AdaRefiner substantially improves the performance, all methods including AdaRefiner fall short in the most difficult level-$8$ task ``Collect Diamond.'' This gap points to a need for further improvements in current methods to tackle more complex tasks.

Nevertheless, the uncovering of knowledge from LLMs by the Adapter LM demonstrates promising prospects for filling the gap in LLMs' performances across various tasks. In future work, we will continue to explore this characteristic of the Adapter LM while also attempting to integrate LLM with RL algorithms more closely to address these limitations in complex environments.

\section*{Ethical Considerations}

While the natural language guidance generated by LLMs exhibits strong common-sense capabilities, there is a possibility that they might contain or produce harmful information. Though no such concerns were observed during evaluations in simulated environments like \textit{Crafter}, it is imperative to address these potential risks when transferring AdaRefiner to more open and real-world settings in the future. Mitigating these risks can be achieved by adding additional instructions in prompts, fine-tuning with curated data, and post-processing the generated text. Adopting these measures ensures that AdaRefiner functions effectively and safely in its intended roles.

\section*{Acknowledgements}
This work was supported in part by NSF China under grant 62250068.

\bibliography{acl_latex}

\clearpage

\appendixpage
\appendix

\section{Pseudo Code for AdaRefiner}\label{sec: pseudo}

\begin{algorithm}[ht]
\caption{Pseudo Code for AdaRefiner}\label{algo:ours}
\begin{algorithmic}[1]
\State \textbf{Init:} Policy $\pi$; Buffer $\mathcal{B}$; Supervised fine-tuning (SFT) buffer $\mathcal{D}$; LLM generation interval $N_\text{gen}$; SFT interval $N_\text{sft}$.
\State $o_0\leftarrow$ \texttt{env.reset}$(),~l_0\leftarrow 0$
\For{$t=0,1,\ldots $}
\State {\color{teal} // generate with Adapter LM and LLM}
\If{$t~\texttt{\%}~N_\text{gen} = 0$}
\State $c_t\leftarrow \mathcal{M}_\text{a}(\texttt{prompt}_\texttt{a}(\mathcal{B}_t, l_t))$
\State $g_t\leftarrow \mathcal{M}_\text{d}(\texttt{prompt}_\texttt{d}(\mathcal{B}_t, c_t))$
\Else
\State $c_t\leftarrow c_{t-1},~g_t\leftarrow g_{t-1}$
\EndIf
\State {\color{teal} // interact with the environment}
\State $a_t\sim \pi(a_t|o_t, f_\text{emb}(g_t))$
\State $o_{t+1}\leftarrow$ \texttt{env.step}$(a_t)$
\State {\color{teal} // update buffer and policy}
\State $\mathcal{B}_{t+1}\leftarrow \mathcal{B}_t\cup (o_t,a_t,o_{t+1},r_t,g_t)$
\State $\pi_{t+1}\leftarrow \texttt{RL\_Update}(\pi_t, \mathcal{B}_{t+1})$
\State {\color{teal} // update SFT buffer}
\State $l_{t+1}\leftarrow \cos(f_\text{emb}(g_t), f_\text{emb}(\tau)),~\tau\sim \mathcal{B}_{t+1}$
% \If{\texttt{ROGUE-L}$(\mathcal{D};[\texttt{prompt}(\mathcal{B}_t,u_{t+1}), c_t])<\theta$}
\State $\mathcal{D}\leftarrow \mathcal{D}\cup [\texttt{prompt}_\texttt{a}(\mathcal{B}_t,l_{t+1}), c_t]$
% \EndIf
\State {\color{teal} // SFT Adapter LM (with interval $N_\text{sft}$)}
\If{$t~\texttt{\%}~N_\text{sft} = 0$}
\State \texttt{SFT}$(\mathcal{M}_\text{a}; \mathcal{D})$
\EndIf
\EndFor
\end{algorithmic}
\end{algorithm}

\section{Implementation Details}\label{ap:details}

\subsection{RL Algorithm}

We use the classic PPO algorithm for policy learning in AdaRefiner, and the hyperparameters are shown in Table \ref{tab:ppo-param}. It is worth noting that AdaRefiner can be flexibly combined with various RL algorithms and is not limited to PPO.

\begin{table}[ht]
\centering
\resizebox{.96\columnwidth}{!}{
\begin{tabular}{@{\hskip 0.25in} c @{\hskip 0.9in} c @{\hskip 0.25in}}
\toprule[1pt]
\textbf{Hyperparameter}      & \textbf{Value}  \\ \midrule[1pt]
policy learning rate    & 7e-4            \\ 
update epoch        & 16              \\ 
$\gamma$                   & 0.97            \\ 
$\varepsilon$ & 1e-8 \\
clip ratio         & 0.1             \\ 
optimizer & \texttt{Adam} \\
\bottomrule[1pt]
\end{tabular}
}
\caption{Hyperparameters for PPO.}
\label{tab:ppo-param}
\end{table}

\newpage

\subsection{Adapter LM}

We use open-source Llama2-7B weight as initial weight for the Adapter LM. In order to reduce computational resources and time consumption, we perform 4-bit quantization on it. The SFT parameters of the Adapter LM are shown in Table \ref{tab:sft-param}.

\begin{table}[ht]
\centering
\resizebox{\columnwidth}{!}{
\begin{tabular}{@{\hskip 0.25in} c @{\hskip 0.9in} c @{\hskip 0.25in}}
\toprule[1pt]
\textbf{Hyperparameter}      & \textbf{Value}  \\ 
\midrule[1pt]
quant type & \texttt{nf4} \\
learning rate & 2e-4\\
batch size    & 4            \\
gradient accumulation step & 1\\
weight decay & 1e-3\\
max grad norm & 0.3\\
warmup ratio & 0.3\\
lora alpha & 16\\
lora dropout & 0.1\\
lora r & 64\\
$N_\text{gen}$ (w/ GPT3.5) & 10\\
$N_\text{gen}$ (w/ GPT4) & 20\\
$N_\text{sft}$ & 1e3\\
\bottomrule[1pt]
\end{tabular}
}
\caption{Hyperparameters for Supervised Fine-Tuning.}
\label{tab:sft-param}
\end{table}

\subsection{Decision LLM}

We call the API interfaces of OpenAI's \texttt{gpt-4} and \texttt{gpt-3.5-turbo} models. The API parameters used are shown in Table \ref{tab:llm-param}.

\begin{table}[ht]
\centering
% \resizebox{.5\columnwidth}{!}{
\begin{tabular}{@{\hskip 0.25in} c @{\hskip 0.9in} c @{\hskip 0.25in}}
\toprule[1pt]
\textbf{Hyperparameter}      & \textbf{Value}  \\ \midrule[1pt]
temperature    & 0.5  \\
top\_p    & 1.0  \\
max\_tokens & 100 \\
\bottomrule[1pt]
\end{tabular}
% }
\caption{Hyperparameters for LLM.}
\label{tab:llm-param}
\end{table}

\subsection{Text Embedding}

For text embedding, we choose the open-source \texttt{paraphrase-MiniLM-L6-v2} model as the encoder.

\clearpage

\begin{table*}[t]
\centering
% \vspace{-2mm}
% \resizebox{\columnwidth}{!}{
\begin{tabular}{cp{4cm}p{4cm}}
\toprule
\textbf{Case} & \textbf{Description} & \textbf{Explanation} \\ 
\midrule
\raisebox{-0.9\totalheight}{\includegraphics[width=0.4\columnwidth]{imgs/case-1.jpg}}
&
The agent tends to place stones between itself and monsters to avoid combat at night (the number of monsters will increase).
& 
Frequent combats are not conducive to maintaining health and can delay other tasks such as resource collection. Therefore, the agent chooses to avoid combat at the appropriate time.
\\ 
\midrule
\raisebox{-0.9\totalheight}{\includegraphics[width=0.4\columnwidth]{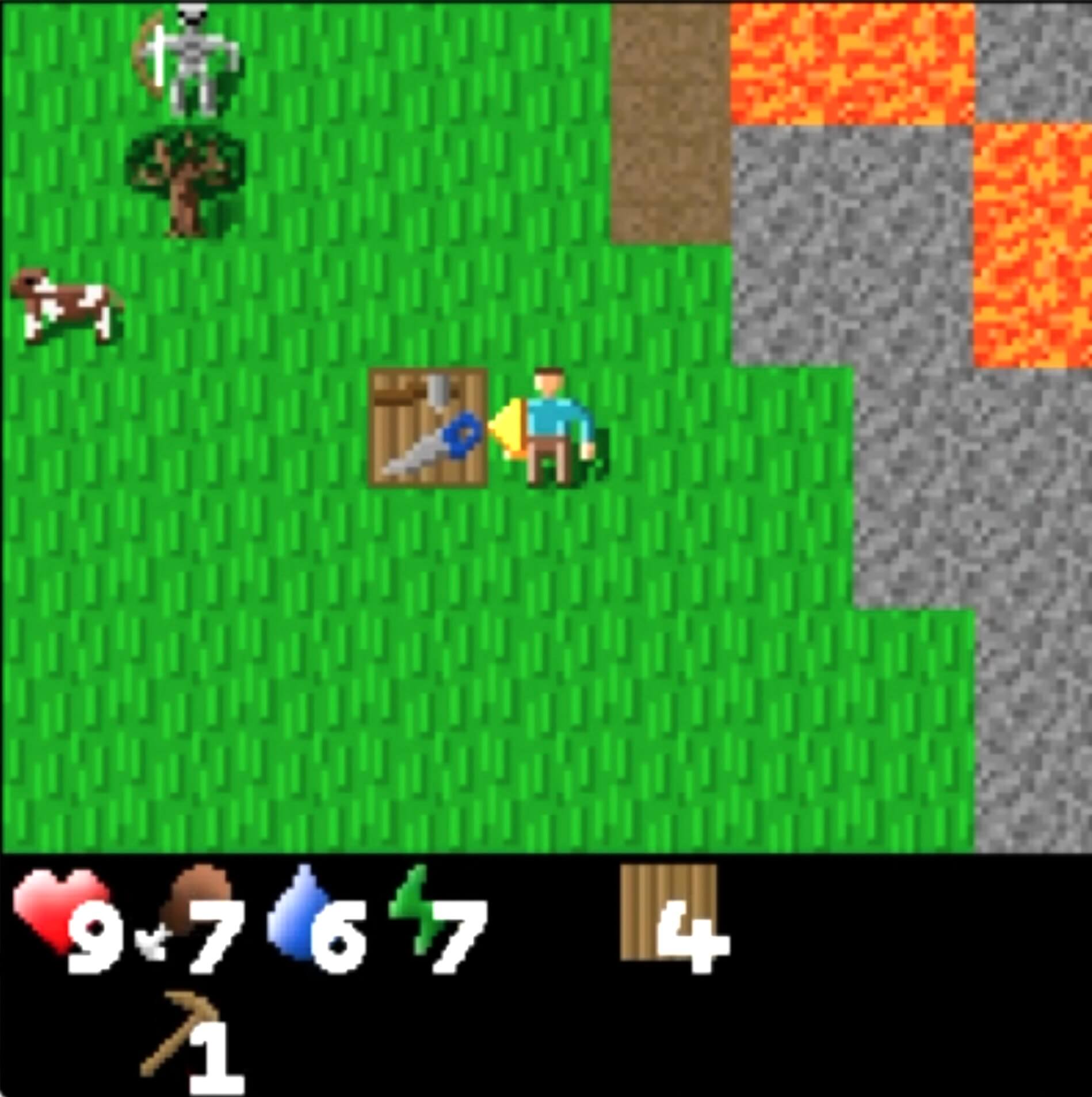}}
&
The agent does not immediately place a workbench to craft tools and unlock achievements when it has abundant resources, but instead places the workbench when moving to resource-rich areas.
& 
Placing the workbench in resource-rich areas can reduce the distance between collecting resources and crafting items, thus improving efficiency.
\\ 
\bottomrule
\end{tabular}
% }
\caption{Case study on agent behaviors grounded in common sense. These behaviors demonstrate the ability of the Adapter LM in uncovering human knowledge behind LLMs.}
\label{tab:cases}
\end{table*}

\begin{figure*}[ht]
    \centering
    \subfloat{
        \includegraphics[width=0.2\textwidth]{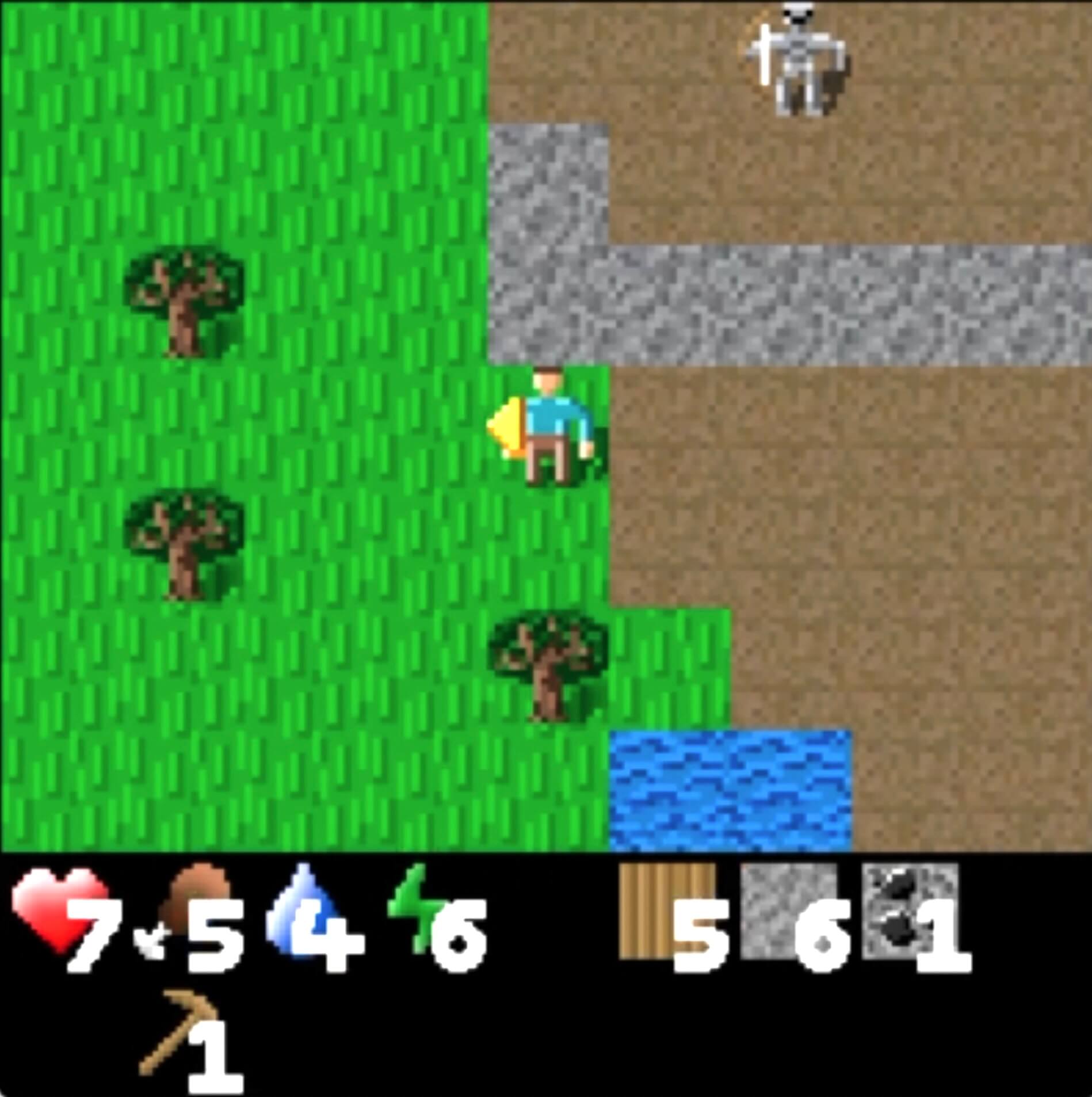}
    }
    \subfloat{
        \includegraphics[width=0.2\textwidth]{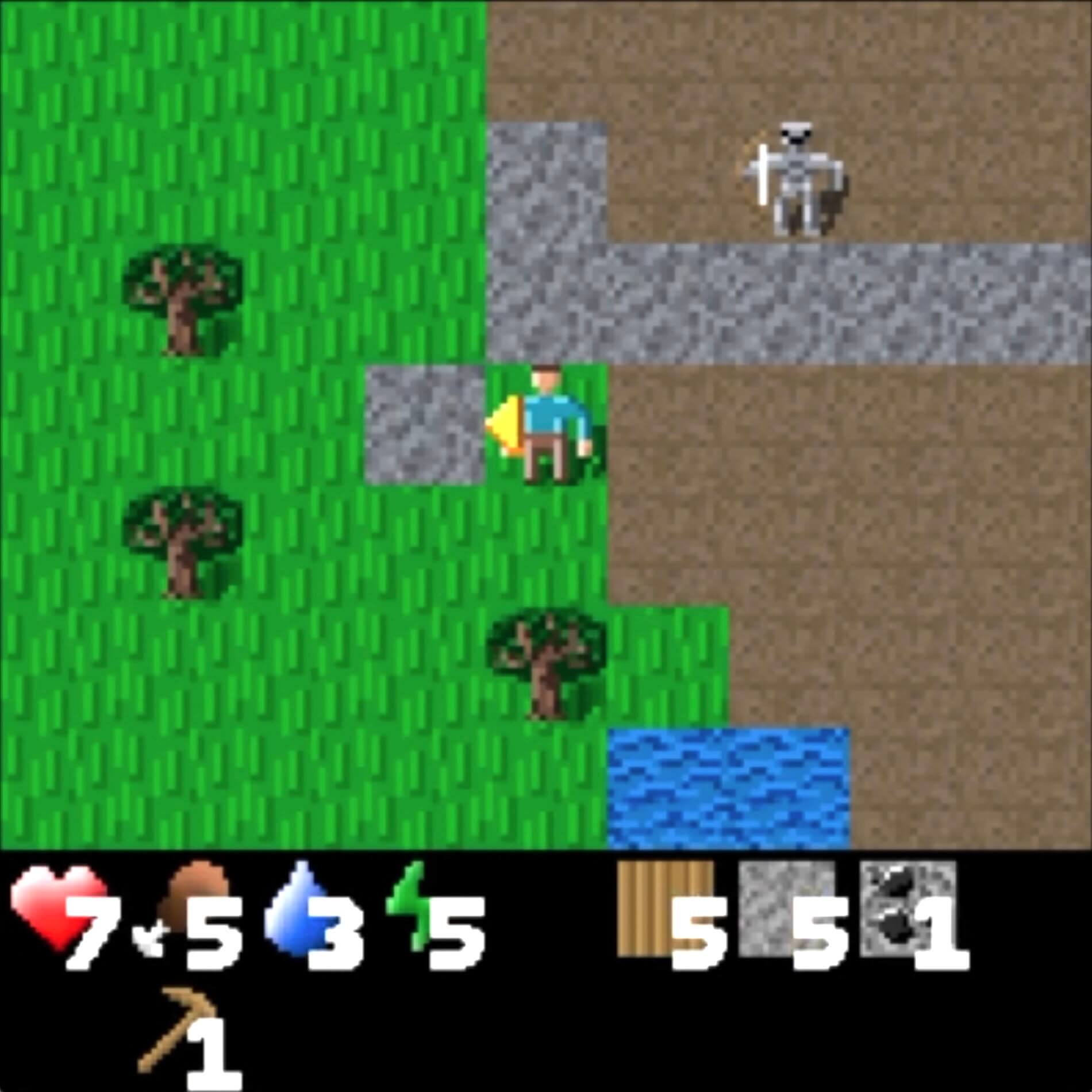}
    }
    \subfloat{
        \includegraphics[width=0.2\textwidth]{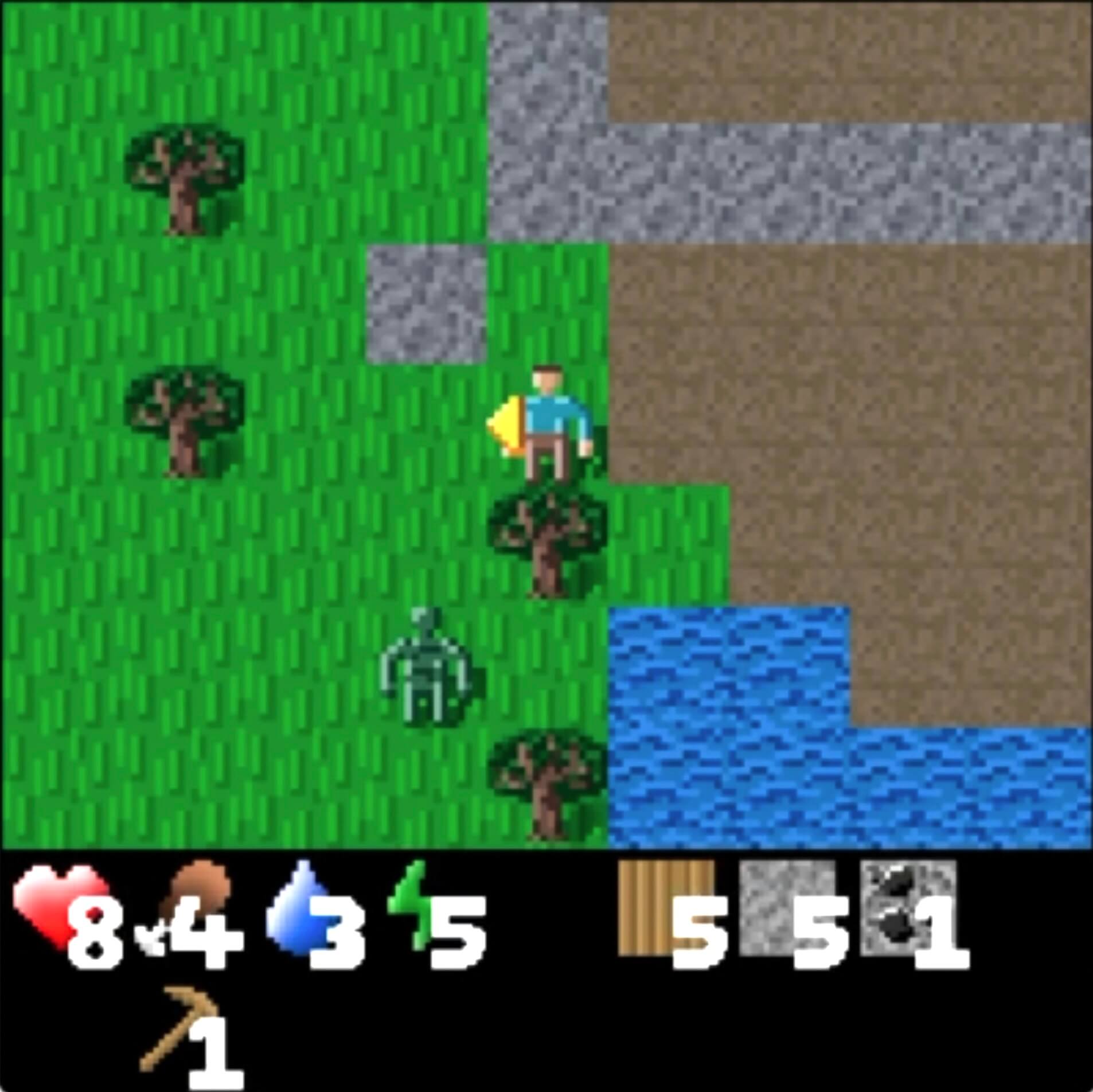}
    }
    \subfloat{
        \includegraphics[width=0.2\textwidth]{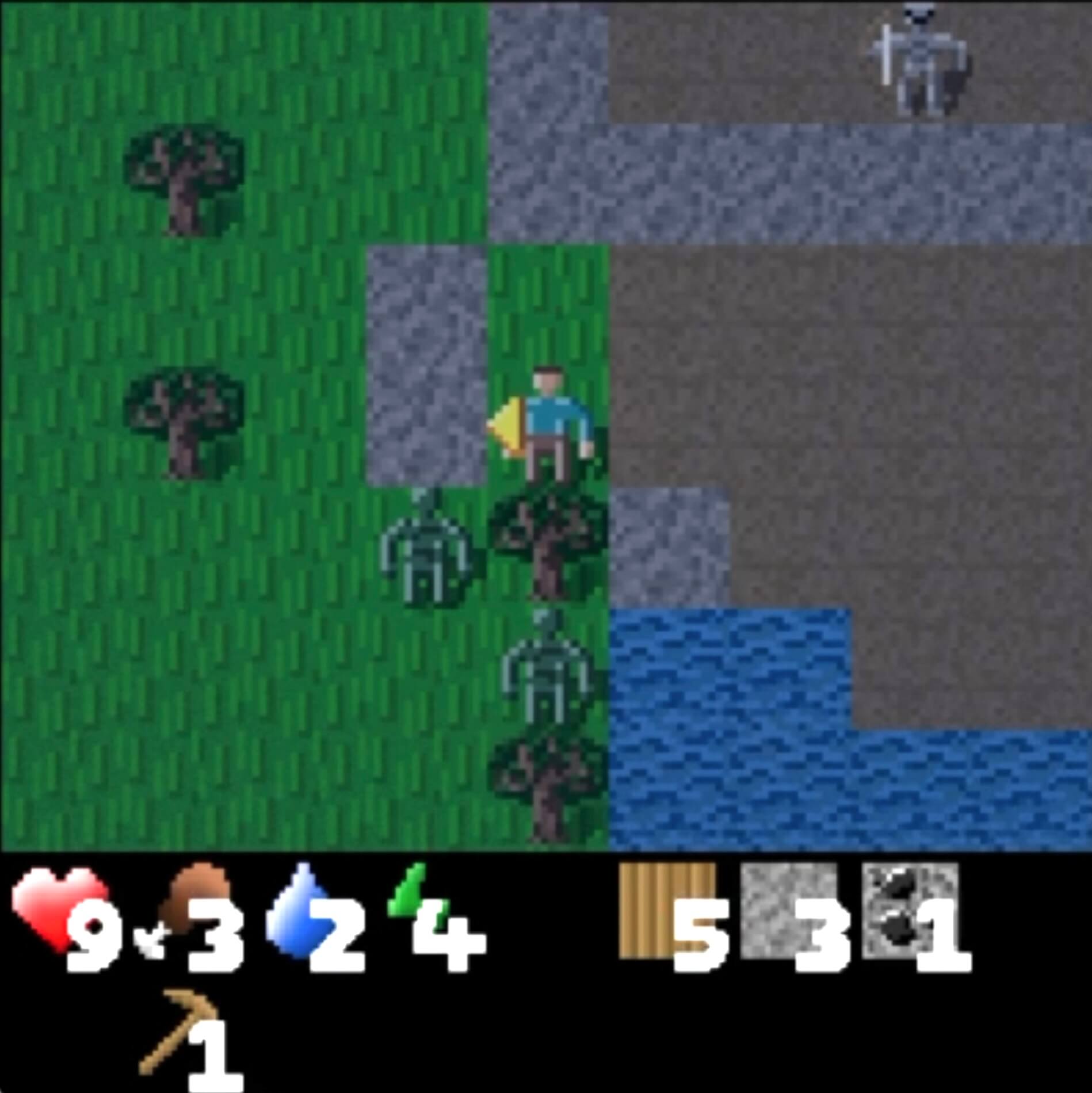}
    }
    \caption{Case details of avoiding combat.}
    \label{fig:case-1-detail}
\end{figure*}

\begin{figure*}[ht]
    \centering
    \subfloat{
        \includegraphics[width=0.2\textwidth]{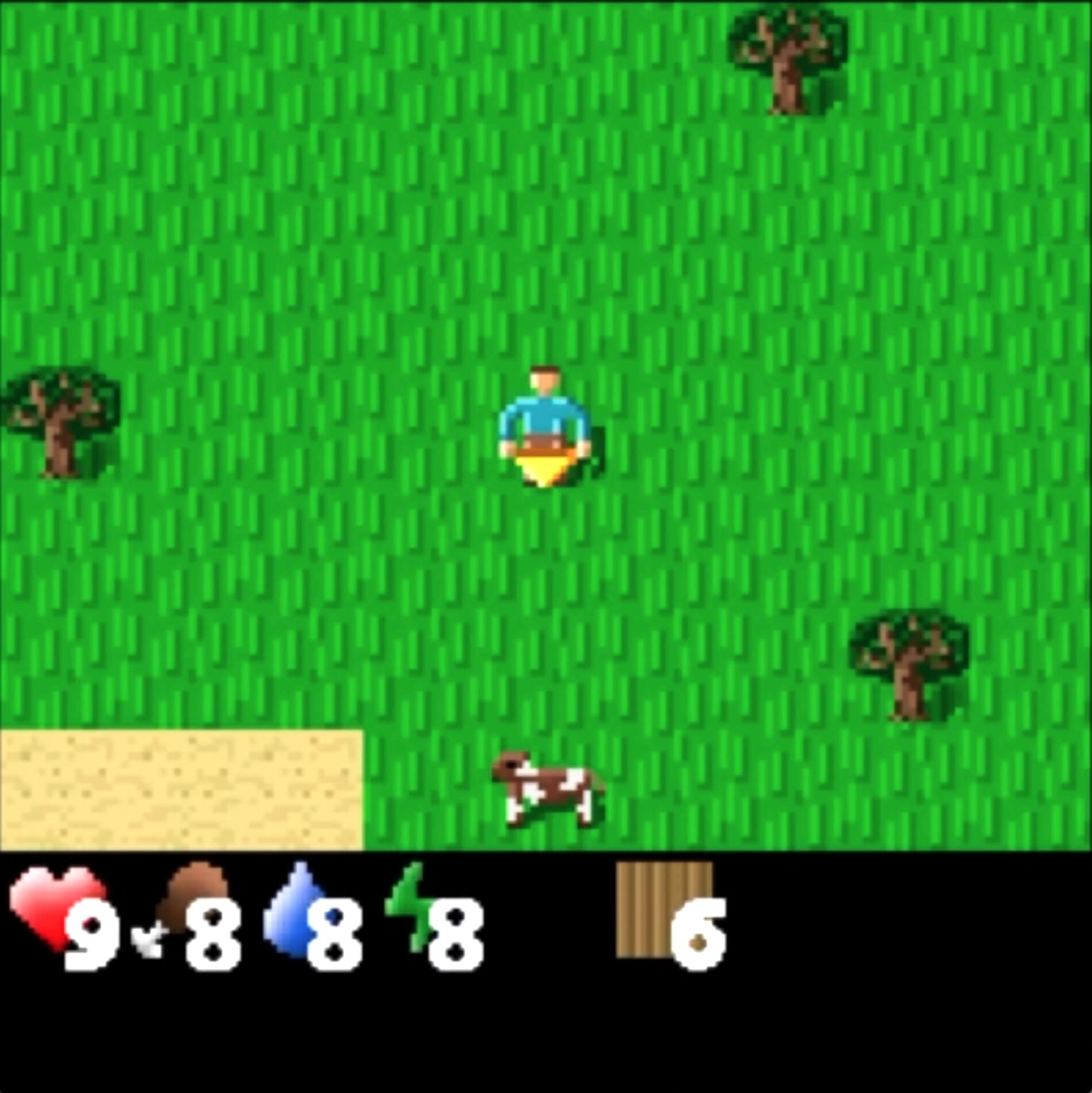}
    }
    \subfloat{
        \includegraphics[width=0.2\textwidth]{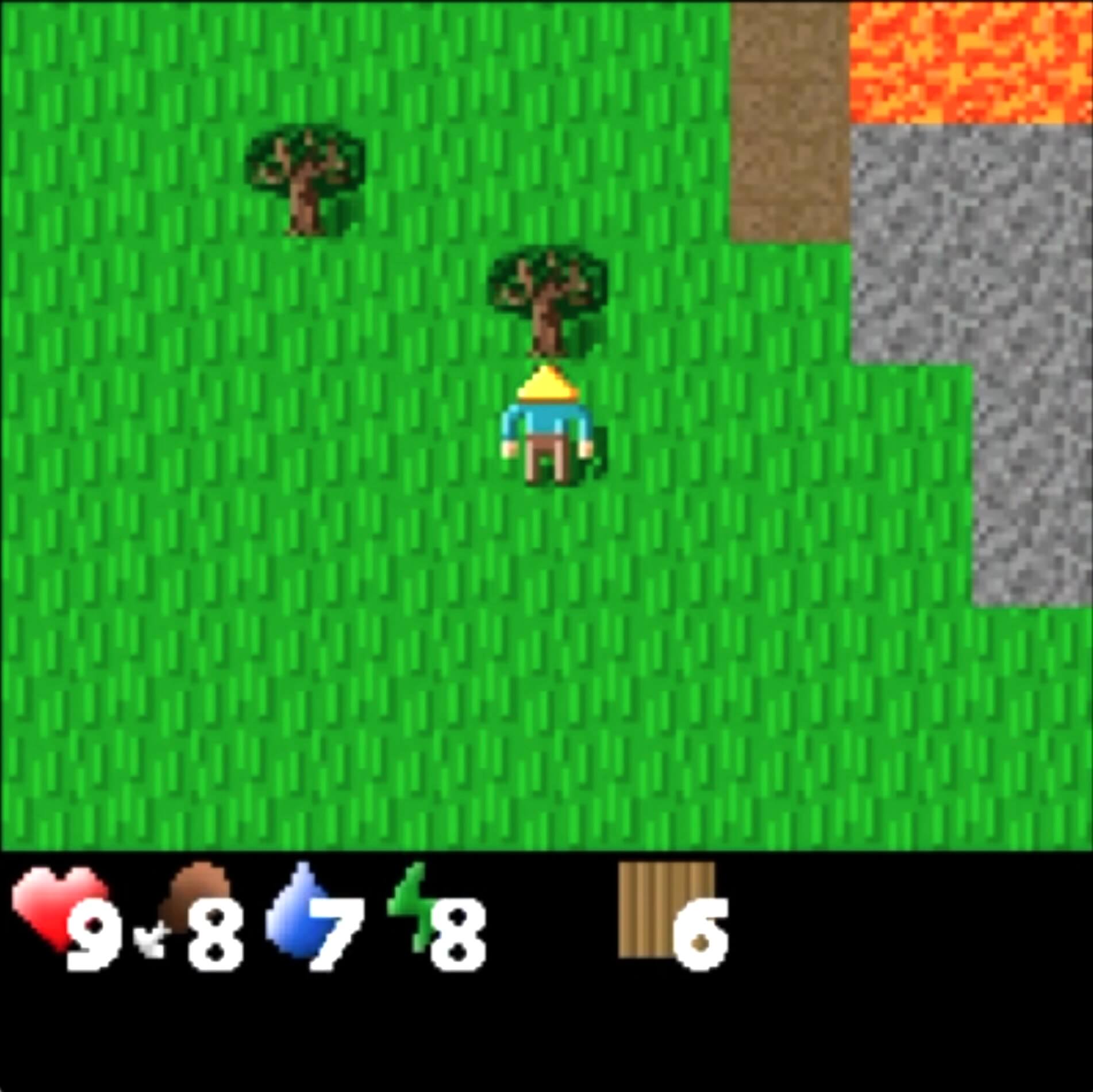}
    }
    \subfloat{
        \includegraphics[width=0.2\textwidth]{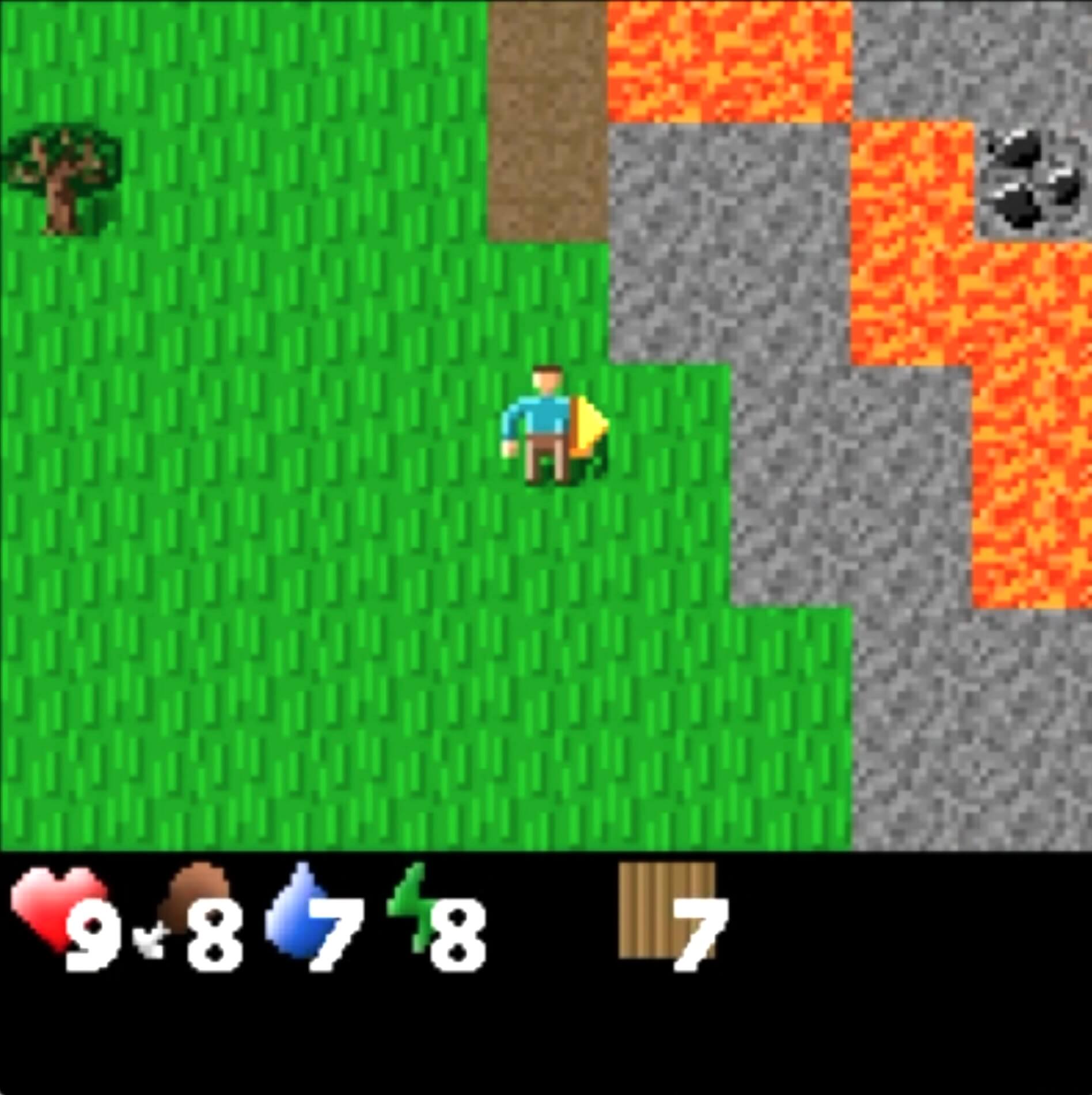}
    }
    \subfloat{
        \includegraphics[width=0.2\textwidth]{imgs/case-2.jpg}
    }
    \caption{Case details of resource planning.}
    \label{fig:case-2-detail}
\end{figure*}

\clearpage

\section{Agent Behaviors Grounded in Common Sense}\label{ap:behaviors}

As discussed in Section \ref{sec:case}, the policy trained by AdaRefiner exhibits behaviors like avoiding combat. Although this may result in a partial performance decrease for the achievements``Defeat Skeleton'' and ``Defeat Zombie'', it could be more advantageous for survival and better completion of other tasks. In this sense, AdaRefiner demonstrates behaviors that align with human common sense. We further analyze additional replays and find other cases of human-like behavior in the policy trained by AdaRefiner, as shown in Table \ref{tab:cases}.

In the two cases, AdaRefiner demonstrates behaviors such as using stones to block monsters and extend survival time, as well as placing workbenches in resource-rich areas for more efficient resource utilization. These behaviors are not observed or reported in other baselines or in the version of AdaRefiner w/o Adapter LM. This further demonstrates that the Adapter LM can better capture the agent's learning ability and uncover common-sense knowledge behind LLMs, prompting them to provide more useful and reasonable guidance for better decision-making.

\subsection{Details of Avoiding Combat}

As shown in Figure \ref{fig:case-1-detail}, it is approaching night and the number of monsters is increasing. The agent starts early to strategically place stones in suitable terrain, successfully building a shelter that can keep the monsters outside and extend its survival time.

\subsection{Details of Resource Planning}

As shown in Figure \ref{fig:case-2-detail}, even though the agent has enough wood to make a workbench, its observations do not reveal abundant resources. Therefore, instead of rushing to make a workbench, it waits until more resources are discovered before making one nearby. This strategy can optimize the efficiency of resource collecting and item crafting.

\section{Full Prompt Details}\label{ap:prompts}

In the following, we provide detailed prompts as well as corresponding example outputs for different methods.

\subsection{Example for Adapter LM}

\texttt{System message:}
\newpage
\begin{lstlisting}
You are a professional game analyst. A player is playing a game similar to Minecraft. Available actions are: 
<move_left, move_right, move_up, move_down, do, sleep, place_stone, place_table, place_furnace, place_plant, make_wood_pickaxe, make_stone_pickaxe, make_iron_pickaxe, make_wood_sword, make_stone_sword, make_iron_sword>. 

You will get the player's observation, status information, and its comprehension score of language guidance (between 0 and 1). You are collaborating with another analyst, and you will be asked to provide concise summaries and suggestions about this player.
\end{lstlisting}

\texttt{Example prompt:}
\begin{lstlisting}
Player sees: <grass, water, cow>

Player status: <7 health, 5 food, 6 drink, 4 energy>

Past action: <sleep>

Past sub-goals:
- eat cow
- collect stone
- place stone

Comprehension score: <0.131>

Analyze the environment and the player's understanding capability,then generate concise summaries and suggestions about this player.
\end{lstlisting}

\texttt{Example output:}
\begin{lstlisting}
The player seems to struggle with understanding past sub-goals, possibly indicating an early stage in the learning process. To help the agent learn the skill of eating cow more quickly, please provide more detailed guidance.
\end{lstlisting}

\subsection{Example for Decsion LLM}

\texttt{System message:}
\begin{lstlisting}
You are a professional game analyst. A player is playing a game similar to Minecraft. Available actions are: 
<move_left, move_right, move_up, move_down, do, sleep, place_stone, place_table, place_furnace, place_plant, make_wood_pickaxe, make_stone_pickaxe, make_iron_pickaxe, make_wood_sword, make_stone_sword, make_iron_sword>. 

You will get analysis about this player from another analyst, and you will be asked to provide the next sub-goals for this player.
\end{lstlisting}

\texttt{Example prompt:}
\begin{lstlisting}
Player sees: <grass, water, cow>

Player status: <7 health, 5 food, 6 drink, 4 energy>

Past action: <sleep>

Past sub-goals:
- eat cow
- collect stone
- place stone

Analysis: <The player seems to struggle with understanding past sub-goals, possibly indicating an early stage in the learning process. To help the agent learn the skill of eating cow more quickly, please provide more detailed guidance.>

Based on the provided information, suggest 3 sub-goals that the player should accomplish next.
\end{lstlisting}

\texttt{Example output:}
\begin{lstlisting}
find cow, move to cow, eat cow
\end{lstlisting}

\subsection{Example for Decsion LLM in \textit{AdaRefiner w/o Adapter LM}}

\texttt{System message:}
\begin{lstlisting}
You are a professional game analyst. A player is playing a game similar to Minecraft. Available actions are: 
<move_left, move_right, move_up, move_down, do, sleep, place_stone, place_table, place_furnace, place_plant, make_wood_pickaxe, make_stone_pickaxe, make_iron_pickaxe, make_wood_sword, make_stone_sword, make_iron_sword>. 

You will get necessary information and player's comprehension score of language guidance (between 0 and 1). You will be asked to provide the next sub-goals for this player.
\end{lstlisting}

\texttt{Example prompt:}
\begin{lstlisting}
Player sees: <grass, tree, water>

Player status: <6 health, 7 food, 3 drink, 1 energy>

Past action: <move_up>

Past sub-goals:
- eat cow
- collect stone
- place stone

Comprehension score: <0.165>

Based on the provided information, suggest 3 sub-goals that the player should accomplish next.
\end{lstlisting}

\texttt{Example output:}
\begin{lstlisting}
collect stone, make stone sword, make stone pickaxe
\end{lstlisting}

\subsection{Example for Vanilla GPT-4}

\texttt{System message:}
\begin{lstlisting}
You are a professional game analyst. A player is playing a game similar to Minecraft. Available actions are: 
<move_left, move_right, move_up, move_down, do, sleep, place_stone, place_table, place_furnace, place_plant, make_wood_pickaxe, make_stone_pickaxe, make_iron_pickaxe, make_wood_sword, make_stone_sword, make_iron_sword>. 

You will get necessary information and player's comprehension score of language guidance (between 0 and 1). You will be asked to provide the next action for this player. You are only allowed to choose the available action.
\end{lstlisting}

\texttt{Example prompt:}
\begin{lstlisting}
You see (object with coordinate)
<cow(2,2), tree(3,1), ...>

Player status: <8 health, 8 food, 8 drink, 6 energy>

Past action: <make_wood_pickaxe>

Comprehension score: <0.275>

Based on the provided information, provide the next action for this player.
\end{lstlisting}

\texttt{Example output:}
\begin{lstlisting}
move_right
\end{lstlisting}

\subsection{Example for Reflexion}

\begin{lstlisting}
Environment: You are playing a game similar to Minecraft. Available actions are 
<move_left, move_right, move_up, move_down, do, sleep, place_stone, place_table, place_furnace, place_plant, make_wood_pickaxe, make_stone_pickaxe, make_iron_pickaxe, make_wood_sword, make_stone_sword, make_iron_sword>. 

Task: Unlock following achievements
<Collect Coal, Collect Diamond, Collect Drink, Collect Iron, Collect Sapling, Collect Stone, Collect Wood, Defeat Skeleton, Defeat Zombie, Eat Cow, Eat Plant, Make Iron Pickaxe, Make Iron Sword, Make Stone Pickaxe, Make Stone Sword, Make Wood Pickaxe, Make Wood Sword, Place Furnace, Place Plant, Place Stone, Place Table, Wake Up>

You see (object with coordinate)
<water(-1,2), cow(2,3), cow(-3,-1), ...>

Your status: 
<9 health, 3 food, 2 drink, 6 energy>
\end{lstlisting}

\begin{lstlisting}
think: Considering the current status, I need to find some food and drink.

think: There are water at (-1,2), I need to move to it.

action: move_left
\end{lstlisting}

\begin{lstlisting}
You see (object with coordinate)
<water(0,2), cow(1,3), cow(-2,-1), ...>

Your status: <9 health, 3 food, 2 drink, 6 energy>
\end{lstlisting}

\begin{lstlisting}
think: Now I move one step to the left, my status has not changed, water is at (0,2). Next, I need to move up.

action: move_up
\end{lstlisting}

\subsection{Example for ReAct}

\begin{lstlisting}
You are playing a game similar to Minecraft. Available actions are 
<move_left, move_right, move_up, move_down, do, sleep, place_stone, place_table, place_furnace, place_plant, make_wood_pickaxe, make_stone_pickaxe, make_iron_pickaxe, make_wood_sword, make_stone_sword, make_iron_sword>. 

Your task is to: Unlock following achievements
<Collect Coal, Collect Diamond, Collect Drink, Collect Iron, Collect Sapling, Collect Stone, Collect Wood, Defeat Skeleton, Defeat Zombie, Eat Cow, Eat Plant, Make Iron Pickaxe, Make Iron Sword, Make Stone Pickaxe, Make Stone Sword, Make Wood Pickaxe, Make Wood Sword, Place Furnace, Place Plant, Place Stone, Place Table, Wake Up>

You see (object with coordinate)
<tree(1,-1), cow(1,4), ...>

Your status: <9 health, 7 food, 8 drink, 7 energy>
\end{lstlisting}

\begin{lstlisting}
think: My status is good, I need to collect some wood to make tools. First I need to move to the tree at (1,-1).

actions: move_right, move_down
\end{lstlisting}

\newpage

\section{Compute Resource Details}

The hardware resources we used are listed in Table \ref{tab:computing-resources}. For each seed, the average GPU running speed is approximately 30K steps/hour.

\begin{table}[ht]
\centering
\resizebox{\columnwidth}{!}{
\begin{tabular}{ccc}
\toprule[1pt]
\textbf{CPU }                     & \textbf{GPU}                      & \textbf{RAM}   \\
\midrule
Intel Xeon 8280@2.7GHz & Nvidia A100 (40GB) & 256GB\\
Intel I9-12900K@3.2GHz & Nvidia RTX 3090 (24GB) & 128GB\\
\bottomrule[1pt]
\end{tabular}
}
\caption{Computational resources for our experiments.}
\label{tab:computing-resources}
\end{table}

\section{Licenses}

In our code, we have used the following libraries which are covered by the corresponding licenses:

\begin{itemize}
    % \item Numpy (BSD-3-Clause license)
    % \item Pandas (BSD-3-Clause license)
    % \item PyTorch (BSD-3-Clause license)
    % \item OpenAI Gym (MIT license)
    \item Crafter (MIT license)
    \item OpenAI GPT (CC BY-NC-SA 4.0 license)
    \item Llama 2 (Llama 2 license)
    \item SentenceTransformer (Apache-2.0 license)
\end{itemize}

\end{document}